\documentclass[11pt]{article}

\usepackage[preprint]{acl}

\usepackage{times}
\usepackage{cuted}
\usepackage{fancyvrb}
\usepackage{latexsym}
\usepackage{tcolorbox}
\usepackage{enumitem}
\tcbuselibrary{breakable}
\usepackage{placeins}
\usepackage{longtable}
\usepackage{booktabs}
\usepackage{graphicx}
\usepackage{subcaption}
\usepackage{amsmath}
\usepackage{amssymb}
\usepackage{array}
\usepackage{booktabs}
\usepackage{tcolorbox}  
\usepackage{fancyvrb}   
\usepackage{fvextra}   
\usepackage{tabularx}   
\usepackage{graphicx}   
\usepackage[table]{xcolor} 
\DeclareMathOperator{\E}{\mathbb{E}}
\tcbuselibrary{listings, skins} 
\usepackage[T1]{fontenc}

\usepackage[utf8]{inputenc}
\usepackage{multicol}
\usepackage{multirow}
\usepackage{microtype}
\usepackage{pifont}
\usepackage{booktabs}
\usepackage{xcolor}
\newcommand{\cmark}{\textcolor{green}{\ding{51}}}%
\newcommand{\xmark}{\textcolor{red}{\ding{55}}}%

\usepackage{inconsolata}

\usepackage{graphicx}
\usepackage{array}
\usepackage{booktabs}
\usepackage{multirow}
\usepackage[table]{xcolor}

\newcommand{\name}{OS-Themis}

\definecolor{lightgray}{gray}{0.9} 
\definecolor{darkgray}{gray}{0.5} 
\definecolor{lightred}{RGB}{255, 120, 120}   
\definecolor{lightgreen}{RGB}{100, 200, 100} 
\definecolor{lightblue}{RGB}{100, 150, 255}  
\definecolor{myblue}{RGB}{219,232,247}
\definecolor{lakeblue}{RGB}{0, 128, 192}
\definecolor{lightyellow}{rgb}{1.0, 1.0, 0.88}  
\definecolor{lightpink}{rgb}{1.0, 0.71, 0.76}   
\definecolor{lightorange}{rgb}{1.0, 0.83, 0.62} 
\definecolor{lightcyan}{rgb}{0.88, 1.0, 1.0}   
\definecolor{lightpurple}{rgb}{0.9, 0.8, 1.0}   
\definecolor{lavender}{rgb}{0.9, 0.8, 1.0}       
\definecolor{lilac}{rgb}{0.78, 0.64, 0.8}        
\definecolor{periwinkle}{rgb}{0.8, 0.8, 1.0}     
\definecolor{mauve}{rgb}{0.87, 0.63, 0.87}       
\definecolor{orchid}{rgb}{0.85, 0.44, 0.84}      
\definecolor{amethyst}{rgb}{0.6, 0.4, 0.8}       
\definecolor{wisteria}{rgb}{0.79, 0.63, 0.86}    
\definecolor{dustylavender}{rgb}{0.76, 0.7, 0.86} 
\definecolor{frenchlavender}{rgb}{0.8, 0.6, 0.7} 
\definecolor{heliotrope}{rgb}{0.87, 0.73, 1.0}   
\definecolor{plum}{rgb}{0.87, 0.63, 0.87}        
\definecolor{LakeBlue}{RGB}{0,61,153}

\definecolor{veronica-red}{RGB}{196,30,58}

\title{\name: A Scalable Critic Framework for Generalist GUI Rewards}

\author{
 \textbf{Zehao Li\textsuperscript{1,2}},
 \textbf{Zhenyu Wu\textsuperscript{2}},
 \textbf{Yibo Zhao\textsuperscript{2}},
 \textbf{Bowen Yang\textsuperscript{2}},
 \textbf{Jingjing Xie\textsuperscript{3}},
 \\
 \textbf{Zhaoyang Liu\textsuperscript{4}},
 \textbf{Zhoumianze Liu\textsuperscript{2}},
 \textbf{Kaiming Jin\textsuperscript{5}},
 \textbf{Jianze Liang\textsuperscript{2}},
 \textbf{Zonglin Li\textsuperscript{2}},
 \\
 \textbf{Feng Wu\textsuperscript{1}},
 \textbf{Bowen Zhou\textsuperscript{2}},
 \textbf{Zun Wang\textsuperscript{2}},
 \textbf{Zichen Ding\textsuperscript{2,\dag}}
\\
\\
 \textsuperscript{1}University of Science and Technology of China,
 \textsuperscript{2}Shanghai AI Laboratory,
 \textsuperscript{3}CUHK MMLab,\\
 \textsuperscript{4}The Hong Kong University of Science and Technology,
 \textsuperscript{5}National University of Singapore
}

\begin{document}
\maketitle
\begin{abstract}
Reinforcement Learning (RL) has the potential to improve the robustness of GUI agents in stochastic environments, yet training is highly sensitive to the quality of the reward function. Existing reward approaches struggle to achieve both scalability and performance. To address this, we propose \name, a scalable and accurate multi-agent critic framework. Unlike a single judge, \name\ decomposes trajectories into verifiable milestones to isolate critical evidence for decision making and employs a review mechanism to strictly audit the evidence chain before making the final verdict.
To facilitate evaluation, we further introduce OmniGUIRewardBench (OGRBench), a holistic cross-platform benchmark for GUI outcome rewards, where all evaluated models achieve their best performance under \name. Extensive experiments on AndroidWorld show that \name\ yields a 10.3\% improvement when used to support online RL training, and a 6.9\% gain when used for trajectory validation and filtering in the self-training loop, highlighting its potential to drive agent evolution. Our code is available at \href{https://github.com/OS-Copilot/OS-Themis}{OS-Copilot/OS-Themis}.
\end{abstract}

\section{Introduction}
Recent advancements in general-purpose Vision-Language Models~\citep{gpt_5,claude_45,comanici2025gemini,Qwen3-VL, wang2025internvl35,guo2025seed1} have fueled the rapid development of GUI agents~\citep{wu2024copilot,agashe2024agent,wang2024gui,nguyen2025gui,yang2026ossymphonyholisticframeworkrobust}. 
While native agents~\citep{wu2024atlas,xu2024aguvis,wang2025opencuaopenfoundationscomputeruse,liu2025scalecua,qin2025uitars} trained on extensive GUI trajectories demonstrate proficiency in digital navigation, they suffer from poor error recovery in stochastic environments, often failing when deviating from expert trajectories~\citep{qin2025uitars,liu2025scalecua,lu2025ui}. This deficiency has precipitated a shift toward reinforcement learning~(RL) to enable adaptive correction~\citep{xu2025mobilerl,lai2025computerrl,wang2025ui}; critically, the success of this paradigm hinges on reliable reward signals to guide policy optimization.

\begin{figure}[t]
    \centering
    \includegraphics[width=0.95\columnwidth]{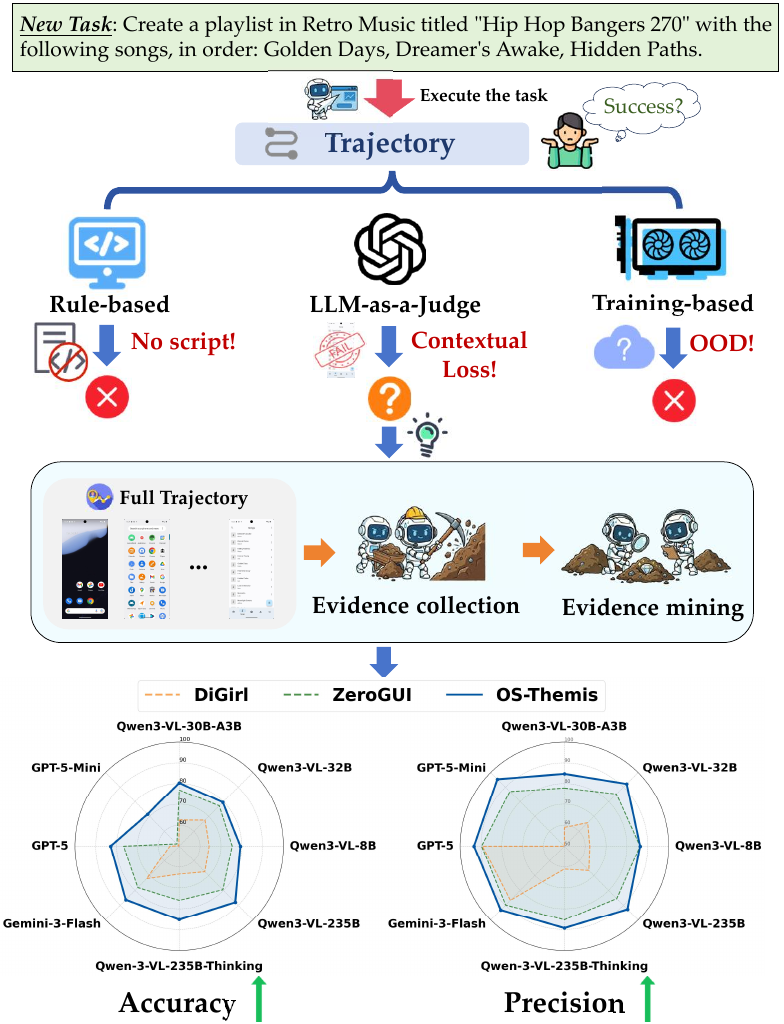} 
    \captionsetup{skip=0pt, position=bottom}
    \caption{Limitations of existing approaches for reward modeling in GUI environments.}
    \label{fig:teaser}
    \vspace{-0.5cm}
\end{figure}
%
As illustrated in Figure~\ref{fig:teaser}, existing methods for reward acquisition in GUI scenarios generally fall into three categories: 1) \textbf{Rule-based rewards} rely on manual heuristics~\citep{ye2025mobile, lai2025computerrl}, offering high precision but suffering from limited scalability and susceptibility to reward hacking. 
2) \textbf{Training-based critics} learn verifiers from human feedback~\citep{qi2024webrl,xu2025mobilerl,wu2025oracle}, yet they demand expensive data construction and generalize poorly to out-of-distribution~(OOD) environments.
3) \textbf{LLM-as-a-judge} offers a flexible and scalable paradigm by harnessing the generalized reasoning capabilities and extensive world knowledge of foundation models~\citep{lee2024prometheus,li2024llms,zhou2025mai}. Despite its potential for generalizable evaluation, existing methods exhibit critical flaws.
First, effective trajectory utilization remains a bottleneck. Sparse sampling~(e.g., last-$K$)~\citep{yang2025zerogui, qi2024webrl, lai2025androidgen} suffers from contextual loss, while global aggregation~\citep{bai2024digirl,wang2025distrl} is compromised by low signal-to-noise ratios in long-horizon tasks. Consequently, both extremes fail to distill decision-critical evidence. 
Second, converting the critical information in trajectories into precise rewards proves challenging. The prevailing \textit{one-shot} paradigm suffers from \textit{evidence dilution}, where accumulated trivial successes mask sparse, outcome-determining failures. This yields overly optimistic judgments that feed wrong-sign signals into Online RL, misleading policy updates.
To address these challenges, we propose \name, a scalable critic framework that shifts from the monolithic single-agent paradigm to a collaborative workflow. 
Specifically: 1) To resolve the trajectory utilization bottleneck, we design a \textbf{Milestone Verification Module} that employs a \textbf{Selector Agent} to decompose long-horizon trajectories into discrete key milestones, assigning an explicit and observable sub-goal to each. Subsequently, a \textbf{Verifier Agent} assesses these milestones sequentially. This granular verification effectively isolates salient signals from irrelevant noise, ensuring precise evidence collection. 
2) To counteract \textit{evidence dilution}, we design the \textbf{Verdict Calibration Module}, consisting of a Reviewer Agent and a Judge Agent. The \textbf{Reviewer Agent} acts as an auditor, meticulously examining the evidence chain prior to final judgment to uncover outcome-determining critical failures masked by trivial successes, thereby correcting overly optimistic assessments and preventing wrong-sign signals.
Based on this structured and audited evidence, \textbf{Judge Agent} renders robust reward signals to enable scalable policy learning.

We conduct extensive experiments to demonstrate the effectiveness of \name. Specifically, we introduce \textbf{OmniGUIRewardBench~(OGRBench)} to mitigate the scarcity of cross-platform outcome reward model~(ORM) benchmarks, where empirical results confirm the state-of-the-art results of \name. In online RL experiments conducted in interactive Android environments, deploying \name\ yields performance gains of up to 7.1\% over baselines on the AndroidWorld benchmark, significantly surpassing competing frameworks. Meanwhile, we conduct a pilot study on scaling RL to validate the effectiveness of \name\ as a reward function, achieving a 10.3\% improvement on AndroidWorld.
 Furthermore, leveraging \name~for trajectory filtering in SFT results in a 6.9\% improvement on the Qwen3-VL series, validating the quality of filtered data and highlighting the framework's potential for autonomous self-evolution.

Our contributions are summarized as follows:

\begin{itemize}[leftmargin=*, itemsep=0pt]
\item We propose \name, a scalable critic framework that delivers reliable reward signals, thereby enabling efficient and robust online RL for GUI agents.
\item We introduce a Milestone Verification and a Verdict Calibration Module to extract critical evidence from GUI trajectories, converting contexts into precise, robust rewards.
\item We construct OGRBench, the first holistic cross-platform ORM benchmark spanning Mobile, Web, and Desktop environments, to comprehensively evaluate GUI agents' critic capabilities.
\item Extensive experiments demonstrate that \name\ significantly boosts online RL performance and facilitates autonomous agent self-evolution in realistic environments.
\end{itemize}

\section{Related Works}

\begin{figure*}[ht]
  \centering
  \includegraphics[width=0.9\linewidth]{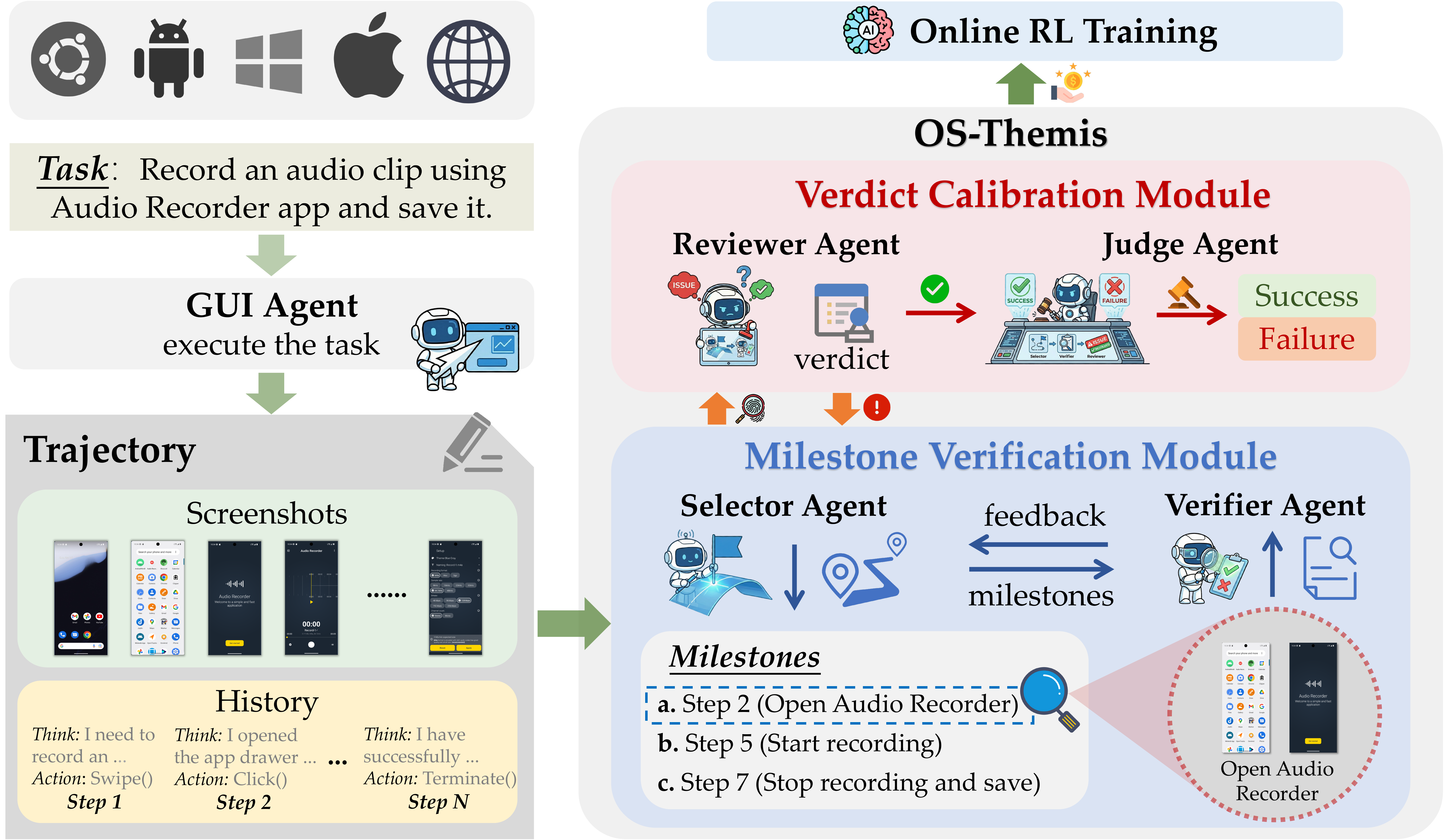}
  \caption{Overview of \name \ Framework. The framework primarily consists of two modules: the \textbf{Milestone Verification Module (MVM)} and the \textbf{Verdict Calibration Module (VCM)}. For a GUI trajectory containing several \texttt{<screenshot,think,action>} steps, the \textbf{Selector Agent} in the \textbf{MVM} extracts key steps as milestones, which are then assigned binary scores by the \textbf{Verifier Agent}. Subsequently, the \textbf{Reviewer Agent} in the \textbf{VCM} continuously interacts with the \textbf{MVM} to ensure the rationality and completeness of the milestones, while the \textbf{Judge Agent} conducts the final scoring of the trajectory based on all information exchanged between the modules.}
  \label{fig:framework}
\end{figure*}

\noindent \textbf{GUI Agents.}
Empowered by general-purpose Vision-Language Models~(VLMs)~\citep{Qwen3-VL,wang2025internvl35,gpt_5,claude_45,comanici2025gemini}, visually-grounded GUI agents have emerged as autonomous copilots for digital tasks~\citep{cheng-etal-2024-seeclick,sun2024genesis,zhang2025breaking,xie2025scalingcomputerusegroundinguser,cua2025,anthropic2024computeruse}.
A dominant paradigm involves \textit{native agents}~\citep{wu2024atlas,qin2025uitars,liu2025scalecua,xu2024aguvis,wang2025opencuaopenfoundationscomputeruse} that integrate planning and grounding in a single-agent architecture, directly translating pixel-level observations into executable actions.
However, despite mastering routine workflows via large-scale training, these agents often exhibit brittleness in stochastic environments, struggling to recover from deviations or generalize to unseen scenarios~\citep{ye2025mobile,liu2025scalecua,lai2025computerrl}.
These limitations have precipitated a shift toward reinforcement learning~(RL) in interactive environments~\citep{xu2025mobilerl,lai2025computerrl,wang2025ui}, where policy optimization hinges on the reliability of reward signals.

\noindent \textbf{Reward Modeling for GUI Agents.}
Reliable reward mechanisms are foundational for GUI agent RL. 
Existing strategies fall into three paradigms:
1) \textbf{Rule-based methods} use heuristic triggers or environment states~\cite{ye2025mobile,lai2025computerrl,wang2025ui}. While providing faithful rewards via verifiable scripts, they scale poorly and are prone to reward hacking.
2) \textbf{Training-based critics} learn verifiers from human feedback or expert trajectories~\cite{qi2024webrl,xu2025mobilerl,wu2025oracle,sun2025seagent,wang2025ui,sun2025sentinel}. Although specialized models like GUI-Critic-R1~\cite{wanyan2025look} and UI-Genie~\cite{xiao2025ui} offer step-wise signals, they struggle with domain shifts and high data collection costs, hindering cross-platform generalization.
3) \textbf{LLM-as-a-judge} harnesses off-the-shelf VLMs for scalable, zero-shot evaluation~\cite{lee2024prometheus,li2024llms,sun2024genesis,murty2024nnetscape,yang2025zerogui,guiactor,zhou2025mai}. Within this paradigm, ZeroGUI~\cite{yang2025zerogui} and DigiRL~\cite{bai2024digirl} represent two classic approaches: sequential evaluation until success versus selecting a fixed number of terminal states.
However, deploying them as Outcome Reward Models~(ORMs) faces critical hurdles: sparse sampling loses critical context~\cite{yang2025zerogui,qi2024webrl}, whereas global ingestion suffers from low signal-to-noise ratios~\cite{bai2024digirl,wang2025distrl}. 
Crucially, bridging this gap requires mobilizing the reasoning capabilities of VLMs to mine fine-grained evidence for precise reward estimation.

\noindent \textbf{GUI Reward Benchmarks.}
Despite the proliferation of benchmarks assessing GUI agent execution~\citep{xie2024osworld,rawles2024androidworld,bonatti2024windows,wang2025mmbenchgui}, datasets evaluating criticism capabilities are scarce. Existing initiatives often suffer from limited scope or data contamination. For instance, GUI-Critic-Test~\cite{wanyan2025look} incurs leakage risks due to its derivation from open-source repositories. Similarly, recent works like OS-Critic Bench~\cite{wu2025oracle}, AgentRewardBench~\cite{men2025agent}, and CUARewardBench~\cite{cua2025} are restricted to isolated domains (e.g., Web or Desktop) or prioritize step-level supervision. 
Consequently, there is an urgent need for a comprehensive ORM evaluation benchmark to establish a reliable evaluation standard for cross-platform reward modeling.

\section{\name}
\subsection{Overview}
In this section, we introduce \name, a multi-agent critic framework for generalist GUI rewards. As shown in Figure~\ref{fig:framework}, the framework primarily consists of two modules: the \textbf{Milestone Verification Module (MVM)} and the \textbf{Verdict Calibration Module (VCM)}. Given a trajectory $\tau$, the MVM first generates an initial milestone set $M_0$, which is then iteratively refined through interaction with the VCM to produce the final milestone set $M$. Subsequently, the VCM assigns a binary reward score $r \in \{0, 1\}$ to the trajectory based on all interaction information between the two modules and the complete refinement process from $M_0$ to $M$.

\subsection{Milestone Verification Module}
Evaluating long-horizon GUI trajectories with a single one-shot judge is prone to errors. Such an approach requires the evaluator to simultaneously identify critical decision points across dozens of steps and reason over the complete interaction history, often leading to overlooked evidence or inconsistent judgments.
Most GUI tasks can be decomposed into a sequence of verifiable subgoals, which we term \emph{milestones}. A milestone represents a necessary intermediate state that must be achieved for task completion, such as "reaching the camera preview" and "capturing the photo" in a photo-taking task. Crucially, verifying whether each milestone is achieved can be done locally by examining the relevant state transition, rather than processing the entire trajectory at once. 

Building on this insight, we introduce the \textbf{Milestone Verification Module (MVM)}, which decomposes trajectories into milestones and verifies them in a structured, step-by-step manner. The module comprises two collaborative agents:

\textbf{Selector Agent.} 
Given a task instruction $\mathcal{I}$ and a trajectory $\tau = \{(s_t, a_t, m_t)\}_{t=1}^T$, where $s_t$ denotes the state (screenshot), $a_t$ the action, and $m_t$ the agent-generated metadata at step $t$, the Selector identifies a compact set of milestone candidates $\mathcal{M}_0 = \{(t_i, d_i, r_i)\}_{i=1}^k$. Each milestone tuple consists of: (i) the step index $t_i$ where verification should occur, (ii) a description $d_i$ of the expected state change or progress, and (iii) a rationale $r_i$ explaining the necessity of this milestone for task completion. The Selector's output provides a structured decomposition that covers all critical subgoals required to judge whether the task has been successfully completed.

\textbf{Verifier Agent.}
For each proposed milestone $(t_i, d_i, r_i) \in \mathcal{M}_0$, the Verifier assesses whether the intended progress described in $d_i$ has been achieved at step $t_i$. It takes as input the pre-action state $s_{t_i}^{\text{pre}}$, post-action state $s_{t_i}^{\text{post}}$, the executed action $a_{t_i}$, and the metadata $m_{t_i}$. The Verifier outputs a binary verification result $v_i \in \{0, 1\}$ indicating milestone achievement, along with grounded feedback $f_i$ when $v_i = 0$. This feedback references specific visual evidence from the screenshots (e.g., "expected confirmation dialog not present," "action targeted incorrect UI element") and provides actionable diagnostic information. The complete verification output can be denoted as $\mathcal{V}_0 = \{(v_i, f_i)\}_{i=1}^k$.

The MVM's milestone-based decomposition offers two key advantages: (1) verification becomes more accurate by focusing on local state transitions rather than holistic trajectory assessment, and (2) failed verifications produce state-grounded feedback that enables systematic refinement of the milestone set, as described in the following section.

\subsection{Verdict Calibration Module}

While the MVM provides a structured approach to trajectory evaluation, the milestone set $\mathcal{M}_0$ generated in a single pass may be incomplete or insufficiently rigorous. The Selector may overlook critical verification points, propose overly lenient success criteria, or fail to capture subtle failure modes, which can in turn lead to false positives. In reinforcement learning, such false positives are particularly detrimental because they reinforce incorrect behaviors. As we show in Appendix~\ref{app:pr_tradeoff}, when recall is already sufficiently high, trading a modest decrease in recall for improved precision yields better gradient quality in policy-gradient-based RL. To address this, we introduce the \textbf{Verdict Calibration Module (VCM)}, which iteratively refines the milestone set through critical review and produces the final reward based on the complete deliberation process. The VCM comprises two agents:

\textbf{Reviewer Agent.}
Given the initial milestone set $\mathcal{M}_0$ and verification results $\mathcal{V}_0$ from the MVM, the Reviewer performs a critical audit to identify whether $\mathcal{M}_0$ is \emph{complete} (covers all necessary subgoals) and \emph{appropriate} (each milestone has clear, verifiable success criteria). Specifically, The Reviewer operates under strict evidence-grounding: any issue it raises must be supported by observable signals in the trajectory. Common concerns include: (i) \emph{missing critical milestones} that leave requirements unverified (e.g., no check for final state persistence), (ii) \emph{overly lenient criteria} that accept intermediate progress without confirming end-state correctness, (iii) \emph{uncaptured failure modes} such as goal-conflicting actions, and (iv) \emph{weak verification evidence} that relies on action descriptions rather than visual confirmation. When issues are identified, the Reviewer generates structured feedback $\mathcal{F} = \{(i_j, q_j)\}_{j=1}^n$, where $i_j$ describes the concern and $q_j$ provides a targeted query. This feedback prompts the MVM to refine $\mathcal{M}_0 \rightarrow \mathcal{M}_1$ and re-verify relevant steps. This iterative process continues until the Reviewer confirms the decomposition is sound, yielding the final $\mathcal{M}$ and $\mathcal{V}$.

\textbf{Judge Agent.}
Once refinement converges or reaches a maximum iteration limit, the Judge synthesizes all information to produce the binary reward $r \in \{0, 1\}$. Crucially, the Judge does not simply aggregate milestone verification results. Instead, it considers the complete deliberation history: (i) milestone evolution $\{\mathcal{M}_0, \ldots, \mathcal{M}\}$, (ii) verification outcomes $\mathcal{V}$, (iii) review feedback and resolution $\mathcal{F}$, and (iv) the original instruction $\mathcal{I}$ and trajectory $\tau$. This enables informed decision-making that goes beyond surface-level verification. For instance, if multiple revision rounds were needed or verification results remain borderline, the Judge weighs these signals toward a more conservative decision. Formally:
\begin{equation}
r = \mathcal{J}\left(\mathcal{I}, \tau, \{\mathcal{M}_0, \ldots, \mathcal{M}\}, \mathcal{V}, \mathcal{F}\right),
\end{equation}
By leveraging the entire refinement process rather than only the final milestone outcomes, this approach substantially reduces false positives while maintaining high recall.

\section{Experiments}

\newcolumntype{N}{>{\centering\arraybackslash}p{2.25em}} 
\newcommand{\gsep}{\hskip 4pt}  
\begin{table*}[!t]
  \centering
  \scriptsize
  \setlength{\tabcolsep}{1.8pt}
  \renewcommand{\arraystretch}{1.1}

  \resizebox{\textwidth}{!}{%
  \begin{tabular}{l
      N N @{\gsep}
      N N @{\gsep}
      N N @{\gsep}
      N N @{\gsep}
      N N @{\gsep}
      N N N N
    }
    \toprule
    \multirow{2}{*}{\textbf{Model}} &
    \multicolumn{2}{c}{\textbf{Ubuntu}} &
    \multicolumn{2}{c}{\textbf{Mobile}} &
    \multicolumn{2}{c}{\textbf{Windows}} &
    \multicolumn{2}{c}{\textbf{macOS}} &
    \multicolumn{2}{c}{\textbf{Web}} &
    \multicolumn{4}{c}{\textbf{Overall}} \\
    \cmidrule(lr){2-3}\cmidrule(lr){4-5}\cmidrule(lr){6-7}\cmidrule(lr){8-9}\cmidrule(lr){10-11}\cmidrule(lr){12-15}
    & \textbf{Acc} & \textbf{F1} &
      \textbf{Acc} & \textbf{F1} &
      \textbf{Acc} & \textbf{F1} &
      \textbf{Acc} & \textbf{F1} &
      \textbf{Acc} & \textbf{F1} &
      \textbf{Acc} & \textbf{Prec} & \textbf{Rec} & \textbf{F1} \\
    \midrule

    \rowcolor{gray!20} \multicolumn{15}{c}{\textbf{DigiRL}} \\
    Qwen3-VL-8B & 62.1 & 64.9 & 69.2 & 76.6 & 66.2 & 63.6 & 72.7 & 40.0 & 63.2 & 69.6 & 64.4 & 62.2 & 72.1 & 66.8 \\
    Qwen3-VL-30B-A3B & 61.0 & 67.6 & 67.0 & 75.6 & 69.0 & 71.1 & 67.5 & 49.0 & 54.2 & 63.0 & 62.5 & 58.9 & \underline{80.6} & 68.1 \\
    Qwen3-VL-32B & 65.7 & 66.5 & 72.3 & 77.8 & 71.8 & 71.2 & 75.3 & 55.8 & 62.1 & 67.9 & 67.6 & 65.8 & 72.1 & 68.8 \\
    Qwen3-VL-235B & 65.5 & 67.0 & 70.2 & 75.2 & 72.3 & 70.9 & 76.6 & 43.8 & 63.2 & 62.8 & 67.4 & 66.6 & 69.0 & 67.8 \\
    Qwen3-VL-235B-Thinking & 62.4 & 66.3 & 67.5 & 74.7 & 65.7 & 66.0 & 70.1 & 46.5 & 58.4 & 62.2 & 63.5 & 61.0 & 73.0 & 66.5 \\
    GPT-5-mini & 47.0 & 0.0 & 47.9 & 0.0 & 55.9 & 0.0 & 79.2 & 0.0 & 47.9 & 0.0 & 50.3 & 0.0 & 0.0 & 0.0 \\
    GPT-5 & 50.5 & 13.7 & 56.9 & 33.1 & 59.1 & 17.1 & 79.2 & 0.0 & 52.6 & 16.7 & 54.5 & 89.3 & 9.6 & 17.3 \\
    Gemini-3-Flash & 69.5 & 63.0 & 69.7 & 64.2 & 76.5 & 66.2 & 83.1 & 31.6 & 75.3 & 74.3 & 72.1 & 86.6 & 51.9 & 64.9 \\
    \rowcolor{gray!10} \textit{Mean} & 60.5 & 51.1 & 65.1 & 59.6 & 67.1 & 53.3 & 75.5 & 33.3 & 59.6 & 52.1 & 62.8 & 61.3 & 53.5 & 52.5 \\

    \midrule

    \rowcolor{gray!20} \multicolumn{15}{c}{\textbf{ZeroGUI}} \\
    Qwen3-VL-8B & 72.7 & 67.8 & 78.7 & 79.6 & 76.5 & 67.5 & 89.6 & 69.2 & 75.8 & 75.0 & 75.4 & 86.3 & 60.1 & 70.9 \\
    Qwen3-VL-30B-A3B & 74.8 & 74.6 & 80.3 & 82.8 & 80.8 & 77.3 & 90.9 & 75.9 & 72.1 & 75.1 & 76.9 & 77.9 & 74.9 & 76.3 \\
    Qwen3-VL-32B & 75.0 & 72.2 & 78.7 & 80.0 & 79.3 & 72.5 & 89.6 & 66.7 & 76.3 & 75.9 & 77.2 & 85.3 & 65.3 & 74.0 \\
    Qwen3-VL-235B & 76.5 & 74.5 & 82.5 & 84.2 & 77.9 & 70.8 & \underline{93.5} & 81.5 & 84.7 & 85.6 & 79.6 & 85.5 & 70.9 & 77.5 \\
    Qwen3-VL-235B-Thinking & 71.8 & 66.9 & 79.8 & 82.1 & 79.3 & 72.5 & 89.6 & 66.7 & 79.0 & 78.3 & 75.9 & 85.3 & 62.3 & 72.0 \\
    GPT-5-mini & 48.5 & 6.4 & 49.5 & 5.9 & 56.3 & 2.1 & 79.2 & 0.0 & 49.0 & 5.8 & 51.5 & 87.0 & 2.9 & 5.5 \\
    GPT-5 & 72.6 & 67.0 & 81.9 & 82.3 & 77.5 & 68.8 & 90.9 & 72.0 & 80.5 & 79.6 & 76.7 & 89.7 & 59.9 & 71.8 \\
    Gemini-3-Flash & 76.1 & 72.8 & 77.1 & 76.0 & 80.8 & 74.5 & \underline{93.5} & 81.5 & 77.4 & 75.1 & 78.1 & 89.8 & 63.0 & 74.1 \\
    \rowcolor{gray!10} \textit{Mean} & 71.0 & 62.8 & 76.1 & 71.6 & 76.0 & 63.2 & 89.6 & 64.2 & 74.4 & 68.8 & 73.9 & 85.8 & 57.4 & 65.3 \\

    \midrule

    \rowcolor{gray!20} \multicolumn{15}{c}{\textbf{\name}} \\
    Qwen3-VL-8B & 77.2 & 75.8 & 85.6 & 85.1 & 72.8 & 61.8 & 85.7 & 59.3 & 86.3 & \underline{87.2} & 79.3 & 86.3 & 69.4 & 77.0 \\
    Qwen3-VL-30B-A3B & 79.5 & 79.4 & 84.6 & 85.0 & 76.5 & 70.9 & 88.3 & 66.7 & 80.5 & 79.3 & 80.3 & 84.7 & 73.7 & 78.8 \\
    Qwen3-VL-32B & 77.6 & 75.2 & 83.5 & 82.1 & 75.1 & 63.5 & 88.3 & 69.0 & 84.7 & 83.3 & 79.6 & 92.2 & 64.3 & 75.8 \\
    Qwen3-VL-235B & \textbf{88.1} & \textbf{88.3} & \textbf{92.3} & \textbf{92.8} & 77.5 & 68.4 & \textbf{94.8} & \textbf{86.7} & \underline{92.1} & \textbf{92.2} & \textbf{88.0} & 92.8 & \textbf{82.3} & \textbf{87.2} \\
    Qwen3-VL-235B-Thinking & 83.4 & 83.7 & 89.4 & 89.6 & \underline{85.5} & \underline{82.5} & \underline{93.5} & \underline{82.8} & 84.7 & 83.4 & 85.2 & 89.3 & 79.9 & 84.3 \\
    GPT-5-mini & 68.8 & 59.7 & 65.4 & 52.5 & 76.5 & 64.3 & 87.0 & 54.5 & 75.8 & 70.5 & 71.5 & \textbf{95.4} & 44.7 & 60.9 \\
    GPT-5 & 82.5 & 81.3 & 80.3 & 78.4 & 84.5 & 80.5 & 88.3 & 60.9 & \textbf{95.9} & 80.4 & 82.9 & \underline{93.4} & 70.6 & 80.4 \\
    Gemini-3-Flash & \underline{85.0} & \underline{84.7} & \underline{91.0} & \underline{91.1} & \textbf{86.9} & \textbf{83.7} & \underline{93.5} & \underline{82.8} & 82.6 & 80.5 & \underline{86.2} & 93.2 & 78.0 & \underline{84.9} \\
    \rowcolor{gray!10} \textit{Mean} & 80.3 & 78.5 & 84.0 & 82.1 & 79.4 & 72.0 & 89.9 & 70.3 & 85.3 & 82.1 & 81.6 & 90.9 & 70.4 & 78.7 \\

    \bottomrule
  \end{tabular}%
  }
  \caption{Performance comparison of different models under the DigiRL, ZeroGUI, and OS-Themis frameworks on OGRBench. Each framework reports Accuracy (Acc) and F1-score (F1), while the Overall performance includes Acc, Precision (Prec), Recall (Rec), and F1. Higher values indicate better performance. \textbf{Bold} and \underline{underlined} denote the best and second-best results, respectively.}
  \label{tab:OmniGUIRewardBench_Simple}
\end{table*}

\subsection{OmniGUIRewardBench}
\paragraph{Benchmark Construction.}
To evaluate the effectiveness of \name\ and its generalizability across heterogeneous platforms, we construct OmniGUIRewardBench (OGRBench), a cross-platform outcome reward model (ORM) benchmark for GUI environments. We compile a dataset of real-world trajectories from five representative benchmarks: AndroidWorld~\citep{rawles2024androidworld}, OSWorld~\citep{xie2024osworld}, WindowsAgentArena~\citep{bonatti2024windows}, macOSArena~\citep{wang2025mmbenchgui}, and WebArena-Lite-v2~\citep{wang2025mmbenchgui}. Each trajectory is represented by the full sequence of screenshots over the entire interaction process, paired with the agent’s model outputs. The trajectory-level outcome label is binary (\texttt{True}/\texttt{False}), indicating whether the overall task is successfully completed. The correctness labels of the trajectories are automatically determined by each benchmark’s built-in evaluation rules.
These trajectories were generated by a diverse suite of GUI agents, including the Qwen3-VL series~(4B, 8B, 235B)~\citep{Qwen3-VL}, UITARS variants~(1.5-7B, 72B-DPO)~\citep{qin2025uitars}, ScaleCUA~(7B, 32B)~\citep{liu2025scalecua}, and Claude-Sonnet-4.5~\citep{claude_45}. 
We employed stratified sampling to ensure broad task coverage while enforcing a balanced class distribution, maintaining a positive sample ratio between 0.45 and 0.55. In total, the resulting evaluation set comprises 1,409 trajectories, consisting of 700 positive and 709 negative samples. More details are provided in Appendix~\ref{ogrbench_details}.
\paragraph{Main Results.}
We evaluate a range of models on OGRBench using \name\ and two primary baselines, DigiRL~\citep{bai2024digirl} and ZeroGUI~\citep{yang2025zerogui}. These two baselines serve as the archetypal paradigms for LLM-as-a-Judge frameworks in GUI reward modeling. As other existing critic frameworks are largely derivative of these foundational approaches, we selected them to ensure a representative comparison. 
ZeroGUI adopts a direct assessment paradigm by feeding last-$K$ states, such as screenshots or structural page information, directly into the model for judgment. In this work, we specifically utilize the final two frames for this baseline, aligning with its original configuration. 
Conversely, DigiRL operates on a sequential verification paradigm. It evaluates states iteratively to determine if the objective is met, and this process continues until a state satisfies the goal or the trajectory terminates. 
Table~\ref{tab:OmniGUIRewardBench_Simple} reports the accuracy across different platforms. We further include precision and recall to provide a granular characterization of performance differences among the frameworks. Across all tested base models, \name\ consistently yields superior results in both accuracy and precision. On average, our method outperforms DigiRL by significant margins, including 18.8\% in accuracy, 29.6\% in precision, 16.9\% in recall, and 26.2\% in F1-score. Similarly, it surpasses ZeroGUI by 7.7\%, 5.1\%, 13.0\%, and 13.4\% across the respective metrics. We provide more detailed evaluation results in Appendix~\ref{more_results}.

\subsection{Online RL}

\paragraph{Online RL Infrastructure.}
To facilitate large-scale parallel trajectory rollouts for online RL training, we established a containerized infrastructure. 
Specifically, each Docker container hosts an independent Android Emulator instance and exposes a remote ADB interface for executing standard GUI operations~(e.g., \texttt{click}, \texttt{swipe}, and \texttt{type}). The system supports real-time screen capture and enforces strict environment isolation; devices are re-initialized before each task to ensure a pristine state. 
This deployment strategy minimizes interference across worker processes, thereby improving the stability and reproducibility of the training phase. Furthermore, the workflow operates asynchronously to maximize efficiency: upon the completion of a trajectory, \name\ is immediately invoked for evaluation and reward calculation.

\paragraph{Task Design.}
Utilizing Qwen3-VL-235B, we automatically synthesize a comprehensive pool of tasks, following the methodology established in \cite{lai2025androidgen}. We then employ a lightweight filtering process to curate a training set of $96$ tasks, while reserving a separate subset of $64$ tasks for validation. Validation primarily relies on the rule-based evaluator to determine success, and uses the reward signal produced by the critic method as an auxiliary monitoring signal.

\paragraph{Training Setup.}
We implement multi-turn online reinforcement learning using the GRPO~\citep{shao2024deepseekmath,guo2025deepseek} algorithm within the Verl framework~\citep{sheng2024hybridflow}.
To prevent over-regularization and encourage extensive exploration, we explicitly disable the KL divergence penalty~(\texttt{disable\_kl=true}, \texttt{kl\_coef=0.0}). Optimization is performed using AdamW~\citep{loshchilov2017decoupled} with a learning rate of $1\times 10^{-6}$, a weight decay of $1\times 10^{-2}$, and a gradient clipping threshold of $1.0$. 
During the rollout phase, we employ a sampling temperature of $1.0$ and generate $n=4$ candidate trajectories per state to enhance exploration coverage. Trajectories are truncated at $50$ steps (\texttt{max\_steps=50}) with a request timeout of $60$ seconds. The training process is conducted over a total of $4$ episodes (\texttt{total\_episodes=4}). We provide additional details of the online RL training in Appendix~\ref{rl_training_details}.


\paragraph{Comparisons.} 
To verify the framework's effectiveness across different scales, we fine-tune two policy backbones: Qwen3-VL-4B and Qwen3-VL-8B. Regarding \name, we instantiate it with two backbone options: Qwen3-VL-8B and Qwen3-VL-235B. We benchmark our approach against two external baselines under identical training configurations: SEAgent~\citep{sun2025seagent}, an open-source critic model, and ZeroGUI, an LLM-as-a-Judge method utilizing Qwen3-VL-235B.
\paragraph{Results on AndroidWorld.}
To systematically evaluate \name\ in the online RL setting, we conducted experiments on the AndroidWorld benchmark (see Table~\ref{tab:RL_AndroidWorld}). For the Qwen3-VL-4B backbone, employing \name\ leads to a 6\% absolute improvement over the baseline, significantly outperforming concurrent methods like ZeroGUI (+5.2\%) and SEAgent (+3.5\%). This performance advantage is even more pronounced in the larger model setting: for Qwen3-VL-8B, fine-tuning with \name\ yields a 7.1\% gain over the baseline, while improving over ZeroGUI and SEAgent by 3\% and 4.7\%, respectively. Notably, the increased gain on the 8B model (7.1\% vs. 6.0\%) suggests that our framework scales effectively, offering greater benefits to larger foundation models.

\begin{table}[t]
  \centering
  \small
  \setlength{\tabcolsep}{4pt}
  \renewcommand{\arraystretch}{1.08}

  \begin{tabularx}{\columnwidth}{@{}lXc@{}}
    \toprule
    \textbf{Backbone} & \textbf{Reward used for RL training} & \textbf{Acc} \\
    \midrule

    \multirow{5}{*}{\textbf{Qwen3-VL-4B}}
      & -- & 45.3 \\
      & SEAgent & 47.8 \\
      & ZeroGUI (Qwen3-VL-235B) & 46.1 \\
      & \name\ (Qwen3-VL-8B) & 50.9 \\
      & \name\ (Qwen3-VL-235B) & \textbf{51.3} \\
    \midrule

    \multirow{5}{*}{\textbf{Qwen3-VL-8B}}
      & -- & 47.6 \\
      & SEAgent & 50.0 \\
      & ZeroGUI (Qwen3-VL-235B) & 51.7 \\
      & \name\ (Qwen3-VL-8B) & 53.4 \\
      & \name\ (Qwen3-VL-235B) & \textbf{54.7} \\
    \bottomrule
  \end{tabularx}

  \caption{RL performance on AndroidWorld with different reward sources. Each row is trained independently from the same initialization for the given backbone.}
  \label{tab:RL_AndroidWorld}
\end{table}

\subsection{Scaling exploration}

\paragraph{Scaling Training Setup.}To evaluate the effectiveness of OS-Themis in a realistic online RL scaling setting, we conduct a pilot study on Qwen3-VL-4B. Specifically, within the \name\ framework, we use Qwen3-VL-235B as the backbone model to score trajectories and provide reward signals. We first use Qwen3-VL-235B to generate a large pool of task templates, filter for suitable ones, and then instantiate 1,024 training tasks from the selected templates. For each task, we roll out 4 trajectories and then optimize the policy with the GRPO algorithm for 1 epoch of online reinforcement learning. To prevent excessive distribution shift from the initial policy during scaling and to stabilize training, we incorporate a KL regularization term, adopting $\texttt{low\_var\_kl}$ with $\texttt{kl\_coef = 0.005}$, so as to balance exploration and robustness.
\begin{figure}[ht]
  \centering
  \includegraphics[width=0.47\textwidth]{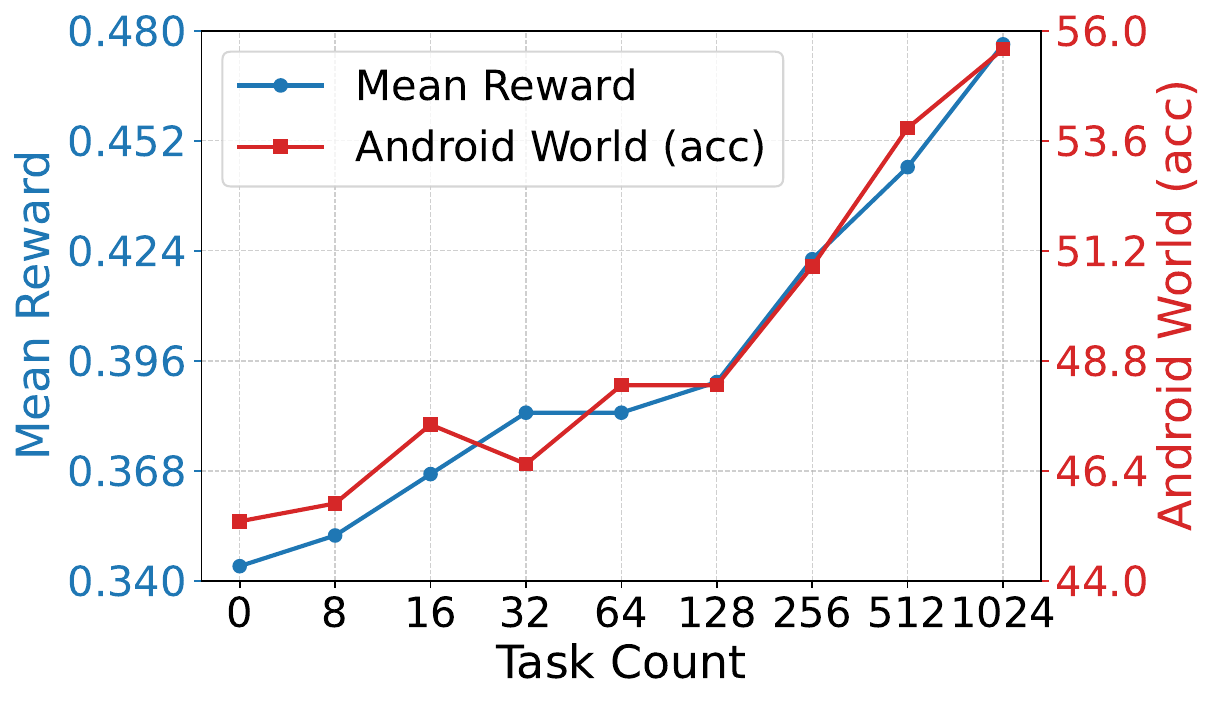}
  \caption{The performance of Qwen3-VL-4B under online RL scaling with \name, including mean reward growth and corresponding AndroidWorld accuracy across different training scales.}
  \label{fig:scaling}
\end{figure}
\paragraph{Scaling Results.}For evaluation, we additionally design 128 extra tasks (disjoint from the 1024 training tasks) as a validation set, and conduct staged validation when the number of training tasks reaches (0, 8, 16, 32, 64, 128, 256, 512, 1024) to characterize the performance trend as the task scale increases. Meanwhile, we evaluate each intermediate model on the AndroidWorld benchmark under the same protocol. As shown in Figure~\ref{fig:scaling}, after scaling training to 1024 tasks, Qwen3-VL-4B achieves a score of 55.6\% on AndroidWorld, improving by 10.3\% over the Baseline. These results demonstrate clear gains and highlight the strong potential of \name\ for scalable online RL.

\paragraph{More Scaling Studies.} In Appendix~\ref{scaling_experiments}, we further conduct a suite of scaling experiments to analyze the scalability of \name\ and its behavior across different usage scenarios. The experiments fall into three categories: \textbf{(1) Model scaling for individual agents}: while keeping the other components at 8B, we separately upgrade the Selector, Reviewer, Judge, and Verifier to Qwen3-VL-235B to quantify each component’s contribution; results suggest that scaling the Judge/Verifier is more critical, while scaling the Reviewer mainly improves precision. \textbf{(2) Framework-level test-time scaling via voting}: we compare three voting strategies (Majority / All / Any) under varying numbers of models ($k$) to characterize controllable precision–recall trade-offs, and observe clear differences across aggregation rules. \textbf{(3) Test-time scaling for evaluation}: On AndroidWorld, we apply a test-time scaling protocol to the policy backbones (Qwen3-VL-4B/8B), where \name\ serves as the online success judge to decide whether to proceed or retry (up to three attempts per task), and the final score is computed by the benchmark’s built-in rules after all attempts finish. For a fair comparison across evaluator frameworks, we instantiate DigiRL, ZeroGUI, and \name\ with the same evaluator backbone, Qwen3-VL-235B. Overall, \name\ yields improved robustness and better evaluation performance.

\subsection{Exploring Self-Evolving Capabilities for GUI Agents}
The advancement of GUI agents is currently constrained by the scarcity of high-quality trajectory data; scaling data acquisition remains a critical bottleneck. However, given that modern agents already possess foundational execution capabilities, a promising solution is to enable autonomous environmental exploration\cite{lai2025androidgen,yan2025stepguitechnicalreport}. This facilitates a virtuous cycle: collected interaction data is used to train the agent, which in turn generates higher-quality data. To sustain this evolution, the primary challenge lies in ensuring the stability and efficiency of data collection while maintaining quality. Effective filtering is therefore essential to distill high-value trajectories from massive exploration logs—a task where \name\ excels. Consequently, we integrate Qwen3-VL and \name\ within a containerized Android environment to establish an autonomous self-evolution paradigm driven by scalable exploration and precise filtering.
\paragraph{Task Generation.}
During the initialization phase, we pre-install a suite of applications within Docker-based Android Emulators and curate a set of representative seed tasks. Utilizing these examples, the model first generates comprehensive descriptions for each application. To ensure a valid exploration context, every session begins with a mandatory app-launch directive; the generation of subsequent exploration tasks is triggered only upon successful application entry. To augment state diversity prior to generation, we inject stochastic interactions (e.g., random swipes or clicks) with a fixed probability. Ultimately, the model synthesizes a series of exploration tasks, conditioned on the real-time device screenshot, the application description, and the seed task examples.

\paragraph{Trajectory Collection and Filtering.}
Upon task generation, the agent initiates execution, interacting with the environment until either the task is completed or a predefined step limit is reached. Crucially, the system operates asynchronously: as soon as a trajectory is concluded, \name\ performs an immediate correctness evaluation, while the collection pipeline concurrently proceeds to the next task. This design enables simultaneous data collection and real-time filtering, efficiently yielding a curated dataset of high-quality trajectories.

\begin{figure}[ht]
  \centering
  \includegraphics[width=0.47\textwidth]{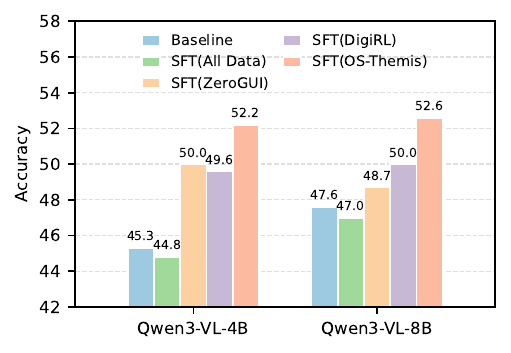}
  \caption{The performance of Qwen3-VL-4B and Qwen3-VL-8B on AndroidWorld after SFT using filtered data (the parentheses indicate the filtering method; All Data means the data is unfiltered).}
  \label{fig:SFT_AndroidWorld}
\end{figure}


\begin{table*}[t]
  \centering
  \small
  \setlength{\tabcolsep}{3.5pt}
  \renewcommand{\arraystretch}{1.05}

  \begin{tabular}{l c c c c N N N}
    \toprule
    \multirow{2}{*}{\textbf{Variant}} &
    \multicolumn{4}{c}{\textbf{Agents}} &
    \multicolumn{3}{c}{\textbf{Overall}} \\
    \cmidrule(lr){2-5} \cmidrule(lr){6-8}
    & \textbf{Selector} & \textbf{Verifier} & \textbf{Reviewer} & \textbf{Judge}
    & \textbf{Acc} & \textbf{Prec} & \textbf{Rec} \\
    \midrule

    \name\ (Full)              & \cmark & \cmark & \cmark & \cmark & \textbf{88.0} & \textbf{92.8} & 82.3 \\
    \name\ w/o Selector        & \xmark & \cmark & \cmark & \cmark   & 83.3         & 79.7         & 88.9 \\
    \name\ w/o Verifier        & \cmark & \xmark & \cmark & \cmark & 81.9         & 77.2         & \textbf{90.1} \\
    \name\ w/o Reviewer        & \cmark & \cmark & \xmark & \cmark & 86.9          & 85.7          & 88.4 \\
    \name\ w/o Judge           & \cmark & \cmark & \cmark & \xmark & 52.5         & 89.7         & 5.0 \\

    \bottomrule
  \end{tabular}

  \caption{Ablation results of OS-Themis by removing individual agents. Metrics include Overall Acc, Prec, and Rec (\%). \textbf{Bold} indicates the best result.}
  \label{tab:ablation}
\end{table*}

\paragraph{Data Quality Evaluation.}
We aggregated a raw dataset of 15,110 trajectories collected in the Android environment. We then filtered the data using DigiRL, ZeroGUI, and OS-Themis, respectively, yielding three high-quality subsets. To evaluate the effectiveness of these filtered data, we conducted supervised fine-tuning (SFT) experiments on the Qwen3-VL-4B and Qwen3-VL-8B backbones. As shown in Fig.~\ref{fig:SFT_AndroidWorld}, fine-tuning on the OS-Themis–filtered data brings substantial improvements of 6.9\% and 5.0\% over the respective baselines, and clearly outperforms fine-tuning on data filtered by DigiRL and ZeroGUI. In contrast, when the entire unfiltered collected dataset is used for SFT, the model performance degrades to varying extents, indicating the presence of substantial noise in the raw data. These results provide strong evidence for the high quality of the filtered data. Beyond this validation, they further indicate that \name\ can serve as a reliable core component for scalable data curation, enabling the construction of an autonomous and self-evolving data collection pipeline.

\subsection{Ablation Study}
\paragraph{Impact of the Selector Agent.} 
In our ablation study, we remove the \textbf{Selector Agent}. Instead of extracting key milestones, we forward every step to the Verifier for validation. While this stepwise design provides denser supervision, it incurs substantially higher verification overhead than the variant that uses the Selector, and its performance on OGRBench is reported in Table~\ref{tab:ablation}.
More importantly, verifying every action introduces a large amount of weakly relevant and noisy evidence, which leads to a pronounced evidence dilution effect: critical failure conditions are obscured by numerous actions that are correct yet nonessential, making it harder to identify the true determinants of success and to attribute errors to outcome-critical decisions. As a result, the downstream judgment becomes overly influenced by accumulated “minor wins,” yielding less discriminative reward signals.
Consequently, compared with the setting that uses the Selector, Accuracy drops by about 4.7\%, and Precision decreases even more sharply by roughly 13.1\%. These results highlight the crucial role of the Selector in mitigating evidence dilution, reducing verification cost, and improving the quality of evidence aggregation for reliable decision making.
\paragraph{Impact of the Verifier Agent.}
We further ablate the \textbf{Verifier Agent} by skipping the verification stage. In this setting, all milestones selected by the Selector Agent are treated as correctly executed by default. After the Selector and Reviewer Agent reach an agreement on the milestone(s), the decision is passed directly to the Judge Agent for final adjudication. The results are summarized in Table~\ref{tab:ablation}. Without the intermediate verification step, the pipeline suffers from systematic bias and reduced reliability, leading to a clear performance degradation: Accuracy drops by 6.1\%, while Precision decreases even more substantially by 15.6\%.

\paragraph{Impact of the Judge Agent.}
We ablate the \textbf{Judge Agent} and determine task success solely based on the correctness of intermediate milestone executions. Concretely, a trajectory is marked as successful only if the Verifier Agent judges all milestones to be correct; otherwise, it is deemed a failure. As shown in Table~\ref{tab:ablation}, this setting leads to a substantial performance drop. In particular, Accuracy decreases to 52.5\%. Although Precision remains high at 89.7\%, Recall collapses to only 5\%, indicating that the system becomes overly conservative in predicting success. This suggests that, for the majority of tasks, occasional imperfect or unsuccessful intermediate operations do not necessarily prevent overall task completion. Therefore, a dedicated Judge Agent is essential to holistically reason over the entire trajectory and infer outcome-level success beyond step-level correctness.
\begin{table}[t]
  \centering
  \small
  \setlength{\tabcolsep}{4pt}
  \renewcommand{\arraystretch}{1.05}
  \begin{tabular}{l N N N}
    \toprule
    \multirow{2}{*}{\textbf{Framework}} &
    \multicolumn{3}{c}{\textbf{Overall}} \\
    \cmidrule(lr){2-4}
    & \textbf{Acc} & \textbf{Prec} & \textbf{Rec} \\
    \midrule
    \name\ w/o Reviewer & 86.9 & 85.7 & \textbf{88.4} \\
    \name\ w/ Reviewer (Advisor) & 87.8 & 88.7 & 86.4 \\
    \name\ w/ Reviewer (Critic) & \textbf{88.0} & \textbf{92.8} & 82.3 \\
    \bottomrule
  \end{tabular}
  \caption{Ablation study on the Reviewer Agent within the OS-Themis framework, comparing versions without a Reviewer and with Reviewers in different roles (Advisor and Critic). Metrics include Overall Acc, Prec, and Rec (\%). \textbf{Bold} indicates the best result.}
  \label{tab:ablation_reviewer}
\end{table}

\paragraph{Impact of the Reviewer Agent.}
To disentangle and quantify the \textbf{Reviewer Agent}'s contribution within \name, we utilize Qwen3-VL-235B as the backbone evaluator and benchmark performance on OGRBench with and without this module. We further explore two distinct role instantiations: \emph{Advisor} and \emph{Critic}. As an \emph{Advisor}, the Reviewer provides constructive suggestions based on traces from the Deep Evaluation Module; as a \emph{Critic}, it rigorously audits the interaction process to uncover overlooked flaws and evidence-grounded failure signals. As detailed in Table~\ref{tab:ablation_reviewer}, incorporating the Reviewer maintains overall Accuracy while substantially boosting Precision by expanding the diversity and coverage of extracted evidence. Although the \emph{Advisor} role yields the most balanced metrics, we ultimately select the \emph{Critic} configuration to better align with the stability requirements of policy-gradient RL. Specifically, we strategically prioritize higher Precision over maximal Recall to minimize false positives, thereby ensuring the high fidelity of reward signals essential for effective policy optimization.

\begin{table}[t]
  \centering
  \small
  \setlength{\tabcolsep}{4pt}
  \renewcommand{\arraystretch}{1.05}

  \begin{tabular}{l N N N}
    \toprule
    \multirow{2}{*}{\textbf{Framework}} &
    \multicolumn{3}{c}{\textbf{Overall}} \\
    \cmidrule(lr){2-4}
    & \textbf{Acc} & \textbf{Prec} & \textbf{Rec} \\
    \midrule
    \name\ w/o Assignment Goal & 86.9 & 84.6 & \textbf{90.0} \\
    \name\ w/ Assignment Goal & \textbf{88.0} & \textbf{92.8} & 82.3 \\
    \bottomrule
  \end{tabular}
  \caption{Ablation study on the Assignment Goal within the OS-Themis framework reporting Overall Acc, Prec, and Rec (\%). \textbf{Bold} indicates the best result.}
  \label{tab:ablation_assignment_goal}
\end{table}

\paragraph{Effectiveness of the Assignment Goal.}
When the Selector Agent proposes a milestone, it concurrently generates an \emph{assignment goal}. This equips the Verifier Agent with a precise, explicit criterion for validation, preventing ungrounded or ambiguous judgments. As demonstrated in Table~\ref{tab:ablation_assignment_goal}, evaluating without an assignment goal leads the Verifier to exhibit excessive leniency, often approving incorrect trajectories. This high false-positive rate degrades overall precision and consequently hampers RL training. Conversely, incorporating assignment goals significantly mitigates this issue, resulting in a marked improvement in model precision.

\section{Conclusion}
In this work, we propose \textbf{\name}, a scalable critic framework designed to mitigate the challenges of contextual information loss and evidence dilution in generalist GUI reward modeling. 
We implement a collaborative workflow where a Milestone Verification Module employs Selector and Verifier agents to isolate salient signals, followed by a Verdict Calibration Module that leverages Reviewer and Judge agents to audit the evidence chain and counteract evidence dilution. This approach fosters the high-precision feedback essential for stable policy optimization. 
Extensive evaluations on our newly proposed OmniGUIRewardBench highlight the superior performance of \name\ across diverse platforms, while empirical results in Online RL and iterative data filtering further underscore its effectiveness in driving capability gains. We hope \name\ serves as a meaningful step toward large-scale reinforcement learning, paving the way for more resilient and self-evolving GUI agents.

\section*{Limitations}
While \name~demonstrates robust initial scalability and stability, we acknowledge certain limitations that delineate promising avenues for future exploration.


\noindent \textbf{Online RL Scaling.} At present, our empirical results mainly demonstrate the feasibility and effectiveness of the proposed framework. However, a systematic characterization of its scaling behavior across task volumes, environment parallelism, training horizons, and model sizes is still constrained by our current RL infrastructure: limited hardware makes it difficult to provision very large numbers of virtual environments, environment coordination and scheduling are not yet sufficiently efficient, and initialization pipelines remain imperfect. With a scalable and efficient online RL training stack, \name\ is expected to deliver more pronounced benefits at larger scales.

\noindent \textbf{Reward Granularity and Formulation}. Although the framework possesses the inherent capacity to provide rich supervision via structured evaluation and fine-grained process extraction, our current exploration of reward shaping remains preliminary. There exists substantial potential to enhance both reward density and learnability. Future work will delve into finer-grained, milestone-wise reward mechanisms and advanced composition strategies. By leveraging process-level evidence to synthesize more stable and information-rich training signals, we aim to further optimize sample efficiency and convergence quality in online optimization.

\section*{Ethical considerations}
The introduction of \name \ for online RL in GUI agents raises ethical considerations primarily related to the stochastic nature of supervision and value alignment. Unlike traditional RL environments defined by deterministic, rule-based success criteria, our framework derives reward signals from the consensus of general-purpose VLMs. This reliance on implicit semantic understanding, rather than explicit programmatic rules, introduces the risk of semantic reward hacking. Without rigid constraints, the policy agent may discover adversarial visual states or actions that exploit the specific reasoning gaps or hallucinations of the VLM critics, triggering high reward scores without genuine task completion.

Furthermore, relying on off-the-shelf VLMs for critique implies inheriting their intrinsic limitations. While the multi-agent architecture aims to mitigate individual errors, there remains a risk of bias propagation where agents collectively reinforce dataset biases found in pre-trained models. This could lead to systematic penalties for legitimate interaction styles in accessibility-oriented interfaces or non-standard GUI layouts that deviate from the VLM's training distribution. Such alignment failures may yield agents whose reliability degrades significantly in diverse, real-world software contexts.

Finally, extending this framework to in-the-wild online training necessitates strict privacy safeguards. Since the agents must continuously process high-fidelity screenshots to compute advantages, the system inevitably handles potentially sensitive user data. To prevent privacy leakage during reward calculation, deployment should remain strictly local, or the system should adopt rigorous data-sanitization protocols, such as removing personally identifiable information~(PII) from visual inputs before model inference. We strictly advocate for a ``Human-in-the-Loop'' validation stage prior to deployment to verify that the agent's learned behaviors align with human intent and have not drifted due to the probabilistic nature of the VLM-based feedback.


\bibliography{custom}
\appendix
\section{Precision-Recall Trade-off in RL}
\label{app:pr_tradeoff}
In policy-gradient-based RL, the reward signal determines \emph{which} trajectories are reinforced. Precision therefore reflects the \emph{purity} of rewarded samples: if many failures are mistakenly rewarded, the update direction is contaminated and the policy may be pushed toward incorrect behaviors, which becomes especially harmful when scaling up training. Recall, by contrast, controls how often truly good behaviors are recognized and rewarded; if recall is too low, positive signal becomes sparse, slowing iteration and limiting the attainable performance ceiling. Our objective is thus to maximize precision while keeping recall sufficiently high so that learning remains effective.

We formalize this trade-off in a simplified setting. Once recall is already adequate to provide stable positive supervision, a mild reduction in recall can be worthwhile if it produces a larger reduction in false positives, thereby increasing precision and improving reward reliability.

Consider a fixed context $x$ where the policy samples a trajectory $\tau\sim\pi_\theta(\cdot\mid x)$ from $\{g,b\}$, with $y(g)=1$ and $y(b)=0$. For compactness, define $\Pr_\theta(\cdot)\triangleq \Pr_{\tau\sim\pi_\theta(\cdot\mid x)}(\cdot)$ and $\E_\theta[\cdot]\triangleq \E_{\tau\sim\pi_\theta(\cdot\mid x)}[\cdot]$. Let
\begin{equation}
\begin{aligned}
p(\theta) \triangleq \Pr_\theta(\tau=g),\quad
1-p(\theta) \triangleq \Pr_\theta(\tau=b).
\end{aligned}
\end{equation}
Assume an imperfect evaluator outputs $\hat r(\tau)\in\{0,1\}$ with operating characteristics
\begin{equation}
\begin{aligned}
\rho  &\triangleq \Pr(\hat r=1\mid \tau=g)\quad\text{(recall / true-positive rate)},\\
\alpha&\triangleq \Pr(\hat r=1\mid \tau=b)\quad\text{(false-positive rate)}.
\end{aligned}
\end{equation}
The true objective is
\begin{equation}
J(\theta) \triangleq \E_\theta[y(\tau)].
\end{equation}
Since $y(g)=1$ and $y(b)=0$, we have
\begin{equation}
J(\theta)=p(\theta).
\end{equation}

The pseudo-objective induced by policy-gradient updates is
\begin{equation}
\hat J(\theta)\triangleq \E_\theta[\hat r(\tau)].
\end{equation}
Expanding by cases yields
\begin{equation}
\hat J(\theta)=p(\theta)\rho+\bigl(1-p(\theta)\bigr)\alpha,
\end{equation}
and equivalently,
\begin{equation}
\hat J(\theta)=\alpha+(\rho-\alpha)\,p(\theta).
\end{equation}
For any baseline $c$ independent of $\tau$,
\begin{equation}
\E_\theta\!\Big[\nabla_\theta\log\pi_\theta(\tau\mid x)\,(\hat r(\tau)-c)\Big]
= \nabla_\theta \hat J(\theta).
\end{equation}
Differentiating gives
\begin{equation}
\nabla_\theta \hat J(\theta)
= \nabla_\theta p(\theta)\,(\rho-\alpha).
\end{equation}
Under the standard logit parameterization $p(\theta)=\sigma(\theta)$,
\begin{equation}
\nabla_\theta p(\theta)=p(\theta)\bigl(1-p(\theta)\bigr)>0,
\end{equation}
so the expected update is governed by the \emph{preference margin} $(\rho-\alpha)$. Intuitively, larger $\alpha$ assigns positive signal to bad trajectories, shrinking the gap between good and bad and reducing reward reliability; when $\alpha$ approaches $\rho$, the reward becomes weakly informative for distinguishing $g$ from $b$.

Precision depends on the policy base rate:
\begin{equation}
\mathrm{Prec}(\theta)\triangleq \Pr(\tau=g\mid \hat r=1).
\end{equation}
By Bayes' rule,
\begin{equation}
\mathrm{Prec}(\theta)
= \frac{p(\theta)\rho}{p(\theta)\rho+\bigl(1-p(\theta)\bigr)\alpha}.
\end{equation}
Maintaining sufficiently high recall $\rho$ ensures true successes are rewarded often enough to provide stable learning signal. However, when $\rho$ is already adequate, reducing the false-positive rate $\alpha$ can increase precision substantially, particularly when $p(\theta)$ is not large (e.g., early training or hard tasks).

Concretely, consider $(\rho,\alpha)\mapsto(\rho',\alpha')$ with $\rho'=\rho-\delta$ and $\alpha'=\alpha-\Delta$, where $\delta>0$ and $\Delta>0$ (locally fixing $p(\theta)$). Then
\begin{equation}
\mathrm{Prec}'(\theta)>\mathrm{Prec}(\theta)
\quad \text{if} \quad
\rho\,\Delta > \alpha\,\delta.
\end{equation}
Meanwhile, the preference margin becomes
\begin{equation}
(\rho'-\alpha')=(\rho-\alpha)+(\Delta-\delta),
\end{equation}
which increases when $\Delta>\delta$, indicating a more reliable discriminative reward for policy-gradient updates.

In summary, once recall is sufficiently high, trading a mild decrease in recall for a larger reduction in false positives can improve precision and yield a more reliable reward signal for policy-gradient-based RL. Nevertheless, recall should not be pushed too low: if $\rho$ becomes small, truly good trajectories are rarely rewarded, leading to sparse positives, higher gradient variance, slower learning, and potentially less stable training.

\begin{table*}[!t]
  \centering
  \scriptsize
  \setlength{\tabcolsep}{1.5pt}
  \renewcommand{\arraystretch}{1.1}

  \caption{Comparison of Accuracy (Acc), Precision (Prec), and Recall (Rec) across various platforms in OmniGUIRewardBench, evaluated using DigiRL, ZeroGUI, OS-Themis, and other GUI critic models (e.g., SEAgent).}
  \label{tab:OmniGUIRewardBench}
  \resizebox{\textwidth}{!}{%
  \begin{tabular}{l ccc ccc ccc ccc ccc ccc}
    \toprule
    \multirow{2}{*}{\textbf{Model}} & \multicolumn{3}{c}{\textbf{Ubuntu}} & \multicolumn{3}{c}{\textbf{Mobile}} & \multicolumn{3}{c}{\textbf{Windows}} & \multicolumn{3}{c}{\textbf{macOS}} & \multicolumn{3}{c}{\textbf{WebA}} & \multicolumn{3}{c}{\textbf{Overall}} \\
    \cmidrule(lr){2-4} \cmidrule(lr){5-7} \cmidrule(lr){8-10} \cmidrule(lr){11-13} \cmidrule(lr){14-16} \cmidrule(lr){17-19}
     & Acc & Prec & Rec & Acc & Prec & Rec & Acc & Prec & Rec & Acc & Prec & Rec & Acc & Prec & Rec & Acc & Prec & Rec \\
    \midrule

    \rowcolor{gray!20} \multicolumn{19}{c}{\textbf{DigiRL}} \\
    Qwen3-VL-8B & 62.1 & 63.7 & 66.2 & 69.2 & 63.3 & \underline{96.9} & 66.2 & 60.6 & 67.0 & 72.7 & 36.8 & 43.8 & 63.2 & 61.1 & 80.8 & 64.4 & 62.2 & 72.1 \\
    Qwen3-VL-30B-A3B & 61.0 & 60.4 & 76.6 & 67.0 & 61.5 & \textbf{98.0} & 69.0 & 60.5 & \textbf{86.2} & 67.5 & 36.4 & \underline{75.0} & 54.2 & 54.4 & 74.8 & 62.5 & 58.9 & \underline{80.6} \\
    Qwen3-VL-32B & 65.7 & 69.0 & 64.1 & 72.3 & 66.9 & 92.9 & 71.8 & 64.9 & \underline{78.7} & 75.3 & 44.4 & \underline{75.0} & 62.1 & 60.8 & 76.8 & 67.6 & 65.8 & 72.1 \\
    Qwen3-VL-235B & 65.5 & 67.9 & 66.2 & 70.2 & 66.4 & 86.7 & 72.3 & 66.1 & 76.6 & 76.6 & 43.8 & 43.8 & 63.2 & 66.3 & 59.6 & 67.4 & 66.6 & 69.0 \\
    Qwen3-VL-235B-Thinking & 62.4 & 63.1 & 70.0 & 67.6 & 62.9 & 91.8 & 65.7 & 58.7 & 75.5 & 70.1 & 37.0 & 62.5 & 58.4 & 59.1 & 65.7 & 63.5 & 61.1 & 73.0 \\
    GPT-5-mini & 47.0 & 0.0 & 0.0 & 47.9 & 0.0 & 0.0 & 55.9 & 0.0 & 0.0 & 79.2 & 0.0 & 0.0 & 47.9 & 0.0 & 0.0 & 50.3 & 0.0 & 0.0 \\
    GPT-5 & 50.5 & 90.6 & 7.4 & 56.9 & 87.0 & 20.4 & 59.2 & 81.8 & 9.6 & 79.2 & 0.0 & 0.0 & 52.6 & \textbf{100.0} & 9.1 & 54.5 & 89.3 & 9.6 \\
    Gemini-3-Flash & 69.5 & 88.5 & 48.9 & 69.7 & 83.6 & 52.0 & 76.5 & 90.7 & 52.1 & 83.1 & \textbf{100.0} & 18.8 & 75.3 & 81.0 & 68.7 & 72.1 & 86.6 & 51.9 \\
    \rowcolor{gray!10} \textit{Mean} & 60.4 & 62.9 & 49.9 & 65.1 & 61.5 & 67.4 & 67.1 & 60.4 & 55.7 & 75.5 & 37.3 & 39.8 & 59.6 & 60.3 & 54.4 & 62.8 & 61.3 & 53.5 \\

    \midrule

    \rowcolor{gray!20} \multicolumn{19}{c}{\textbf{ZeroGUI}} \\
    Qwen3-VL-8B & 72.7 & 90.6 & 54.2 & 78.7 & 79.6 & 79.6 & 76.5 & 86.7 & 55.3 & 89.6 & 90.0 & 56.3 & 75.8 & 81.2 & 69.7 & 75.4 & 86.3 & 60.1 \\
    Qwen3-VL-30B-A3B & 74.8 & 80.1 & 69.7 & 80.3 & 76.1 & 90.8 & 80.8 & 80.5 & 74.5 & 90.9 & 84.6 & 68.8 & 72.1 & 70.2 & 80.8 & 76.9 & 77.9 & 74.9 \\
    Qwen3-VL-32B & 75.0 & 88.2 & 61.1 & 78.7 & 78.4 & 81.6 & 79.3 & 87.9 & 61.7 & 89.6 & \textbf{100.0} & 50.0 & 76.3 & 80.7 & 71.7 & 77.2 & 85.3 & 65.3 \\
    Qwen3-VL-235B & 76.5 & 87.9 & 64.6 & 82.5 & 79.3 & 89.8 & 77.9 & 85.1 & 60.6 & \underline{93.5} & \textbf{100.0} & 68.8 & 84.7 & 84.3 & 86.9 & 79.6 & 85.5 & 70.9 \\
    Qwen3-VL-235B-Thinking & 71.8 & 88.7 & 53.7 & 79.8 & 76.3 & 88.8 & 79.3 & 87.9 & 61.7 & 89.6 & \textbf{100.0} & 50.0 & 79.0 & 84.7 & 72.7 & 75.9 & 85.3 & 62.3 \\
    GPT-5-mini & 48.5 & 86.7 & 3.3 & 49.5 & \textbf{100.0} & 3.1 & 56.3 & \textbf{100.0} & 1.1 & 79.2 & 0.0 & 0.0 & 49.0 & 75.0 & 3.0 & 51.5 & 87.0 & 2.9 \\
    GPT-5 & 72.6 & 92.8 & 52.4 & 81.9 & 84.0 & 80.6 & 77.5 & 88.3 & 56.4 & 90.9 & \textbf{100.0} & 56.3 & 80.5 & 87.8 & 72.7 & 76.7 & 89.7 & 59.9 \\
    Gemini-3-Flash & 76.1 & 91.9 & 60.3 & 77.1 & 84.0 & 69.4 & 80.8 & 89.6 & 63.8 & \underline{93.5} & \textbf{100.0} & 68.8 & 77.4 & 87.8 & 65.7 & 78.1 & 89.8 & 63.0 \\
    \rowcolor{gray!10} \textit{Mean} & 71.0 & 88.4 & 52.4 & 76.1 & 82.2 & 73.0 & 76.1 & 88.2 & 54.4 & 89.6 & 84.3 & 52.3 & 74.3 & 81.5 & 65.4 & 73.9 & 85.8 & 57.4 \\

    \midrule

    \rowcolor{gray!20} \multicolumn{19}{c}{\textbf{OS-Themis}} \\
    Qwen3-VL-8B & 77.2 & 86.6 & 67.4 & 85.6 & 92.8 & 78.6 & 72.8 & 81.0 & 50.0 & 85.7 & 72.7 & 50.0 & 86.3 & 84.8 & \textbf{89.9} & 79.4 & 86.3 & 69.4 \\
    Qwen3-VL-30B-A3B & 79.5 & 84.9 & 74.6 & 84.6 & 86.3 & 83.7 & 76.5 & 78.2 & 64.9 & 88.3 & 81.8 & 56.3 & 80.5 & 88.8 & 71.7 & 80.3 & 84.7 & 73.7 \\
    Qwen3-VL-32B & 77.6 & 91.3 & 63.9 & 83.5 & \underline{94.7} & 72.5 & 75.1 & 90.2 & 48.9 & 88.3 & 76.9 & 62.5 & 84.7 & \underline{97.3} & 72.7 & 79.6 & 92.2 & 64.3 \\
    Qwen3-VL-235B & \textbf{88.1} & 92.2 & \textbf{84.7} & \textbf{92.3} & 93.8 & 91.8 & 77.5 & 89.7 & 55.3 & \textbf{94.8} & \underline{92.9} & \textbf{81.3} & \underline{92.1} & 95.7 & \underline{88.9} & \textbf{88.0} & 92.8 & \textbf{82.3} \\
    Qwen3-VL-235B-Thinking & 83.4 & 87.5 & \underline{80.2} & 89.4 & 91.5 & 87.8 & \underline{85.5} & 88.0 & 77.7 & \underline{93.5} & 92.3 & \underline{75.0} & 84.7 & 96.1 & 73.7 & 85.2 & 89.3 & 79.9 \\
    GPT-5-mini & 68.8 & \textbf{95.0} & 43.5 & 65.4 & 92.3 & 36.7 & 76.5 & \underline{97.8} & 47.9 & 87.0 & \textbf{100.0} & 37.5 & 75.8 & 96.5 & 55.6 & 71.5 & \textbf{95.4} & 44.7 \\
    GPT-5 & 82.5 & \underline{93.7} & 71.8 & 80.3 & 91.8 & 68.4 & 84.5 & 90.7 & 72.3 & 88.3 & \textbf{100.0} & 43.8 & \textbf{95.9} & 70.7 & 81.4 & 82.9 & \underline{93.4} & 70.6 \\
    Gemini-3-Flash & \underline{85.0} & 92.5 & 78.1 & \underline{91.0} & 93.6 & 88.8 & \textbf{86.9} & 92.3 & 76.6 & \underline{93.5} & 92.3 & \underline{75.0} & 82.6 & 97.1 & 68.7 & \underline{86.2} & 93.2 & 78.0 \\
    \rowcolor{gray!10} \textit{Mean} & 80.3 & 90.5 & 70.5 & 84.0 & 92.1 & 76.0 & 79.4 & 88.5 & 61.7 & 89.9 & 88.6 & 60.2 & 85.3 & 90.9 & 75.3 & 81.6 & 90.9 & 70.4 \\
    
    \midrule

    \rowcolor{gray!20} \multicolumn{19}{c}{\textbf{Other Critic Model}} \\
    SEAgent & 59.4 & 94.2 & 24.9 & 55.9 & 85.7 & 18.4 & 63.9 & 87.0 & 21.3 & 84.4 & 100.0 & 25.0 & 65.8 & 82.7 & 43.4 & 61.8 & 89.7 & 26.1 \\
    
    \bottomrule
  \end{tabular}
  }
\end{table*}

\section{More Results}
\label{more_results}
\subsection{Detailed OGRBench results}
In Table~\ref{tab:OmniGUIRewardBench}, we report more detailed, platform-specific metrics for evaluating OGRBench across different models. By analyzing how each framework performs on individual platforms, we observe that, compared with ZeroGUI\cite{yang2025zerogui} and DigiRL\cite{bai2024digirl}, \name\ exhibits consistently strong performance with superior stability and overall effectiveness across diverse platforms.

We also attempted to evaluate several training-based GUI critic models; however, some existing critics are PRM-style models and are not compatible with OGRBench evaluation\cite{wu2025oracle,wanyan2025look,gu2025ui}. For ORM-style critics, most existing models are not open-sourced\cite{xu2025mobilerl,ye2025mobile}, and the few open-source ones are further limited by platform-specific information\cite{qi2024webrl} and thus fail to generalize. Therefore, we only benchmark SEAgent\cite{sun2025seagent} as a representative comparator, but its performance is also not satisfactory.

The detailed results suggest that overall framework performance is strongly influenced by the capability of the underlying model. For instance, GPT-5 and GPT-5-mini\cite{gpt_5} tend to produce more conservative outputs and prefer predicting \texttt{False}, which can lead to notably low recall across different frameworks; \name\ partially mitigates this issue by reducing the extent to which recall is dominated by conservative bias. Meanwhile, Gemini-3-Flash\cite{comanici2025gemini} achieves the highest precision across all models while maintaining sufficiently high recall, indicating more reliable positive judgments.

For the Qwen3-VL\cite{Qwen3-VL} series, we observe that as model size increases, the advantage of \name\ over DigiRL and ZeroGUI becomes more pronounced. With smaller models, the gap in precision is relatively large, whereas the difference in accuracy is limited; however, as reasoning capability improves with scale, the gains brought by \name\ expand substantially, yielding clearer performance separation. This trend suggests that under more complex and evidence-heavy settings, the quality of evidence organization and aggregation becomes a key determinant of final performance, and stronger models can better leverage the structured advantages of \name.

\begin{figure}[t]
  \centering
  \includegraphics[width=0.45\textwidth]{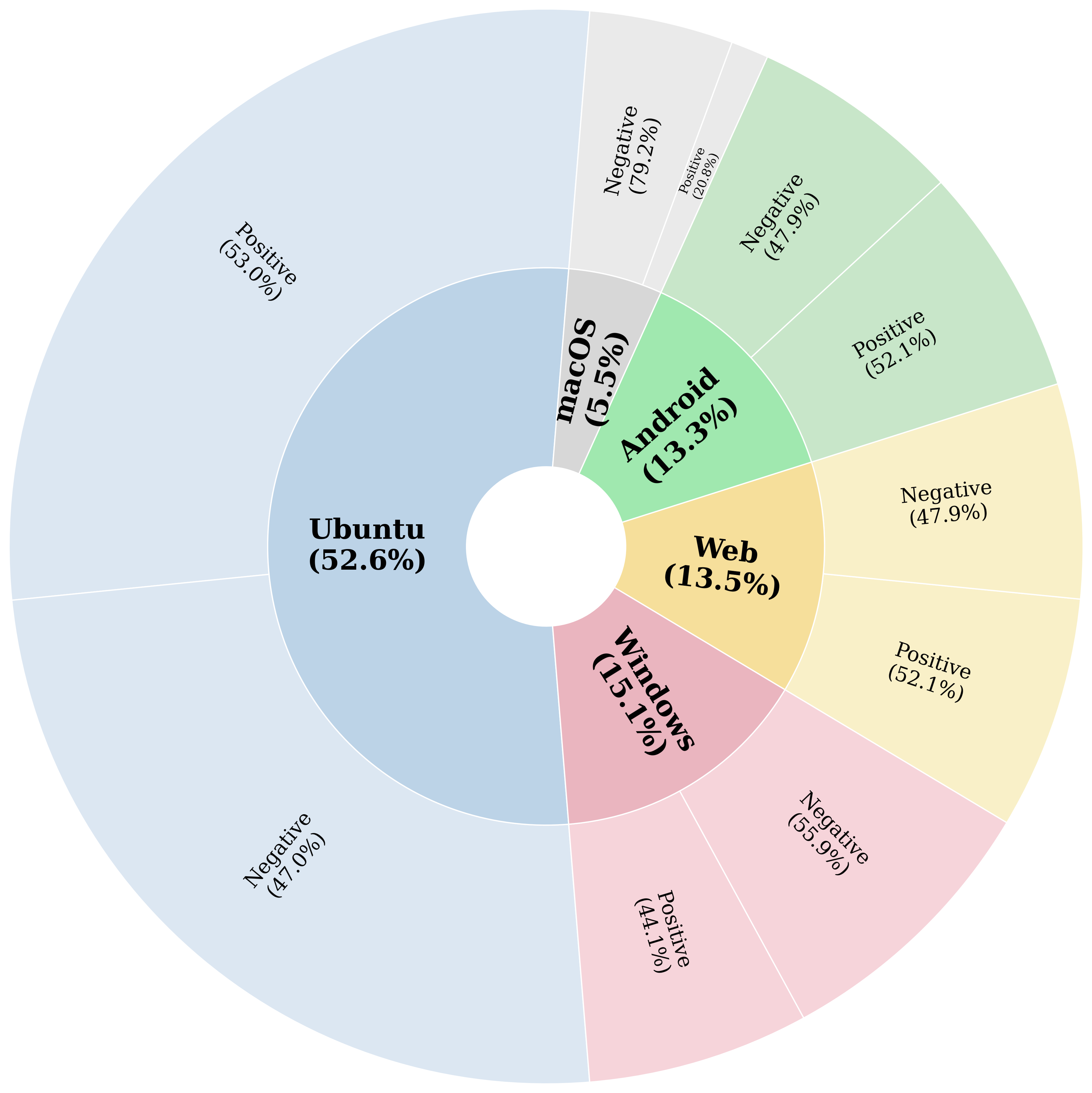}
  \caption{Data Distribution of OGRBench}
  \label{fig:platform_distribution}
\end{figure}
\begin{table}[t]
  \centering
  \small
  \setlength{\tabcolsep}{3pt} 
  \renewcommand{\arraystretch}{1.05}

  \resizebox{\linewidth}{!}{%
  \begin{tabular}{l l N N}
    \toprule
    \multirow{2}{*}{\textbf{Category}} &
    \multirow{2}{*}{\textbf{Method}} &
    \multicolumn{2}{c}{\textbf{Overall}} \\
    \cmidrule(lr){3-4}
    & & \textbf{Prec} & \textbf{Recall} \\
    \midrule

    Official      & Rule-based         & 83.8 & 55.9 \\
    \midrule

    Critic Model  & SEAgent            & 71.6 & {-}  \\
    \midrule

    \multirow{7}{*}{LLM-as-a-Judge}
                 & Claude-3.7-Sonnet   & 64.3 & 89.8 \\
                 & GPT-4o              & 68.1 & 80.3 \\
                 & GPT-4o Mini         & 64.5 & 78.3 \\
                 & Qwen2.5-VL-72B          & 64.5 & 86.1 \\
                 & Qwen3-VL-8B          & 66.3 & 82.0 \\
                 & Qwen3-VL-32B          & 72.9 & 73.9 \\
                 & Qwen3-VL-235B          & 67.5 & 81.7 \\
    \midrule

    \multirow{3}{*}{\name}
                 & Qwen3-VL-8B         & 71.6 & 39.0 \\
                 & Qwen3-VL-32B        & 76.8 & 39.3 \\
                 & Qwen3-VL-235B       & 76.1 & 54.4 \\
    \bottomrule
  \end{tabular}%
  }
  \caption{Performance of different models on AgentRewardBench within the OS-Themis framework.}
  \label{tab:agent_reward_bench}
\end{table}

\subsection{AgentRewardBench}
We evaluated \name\ on AgentRewardBench\cite{men2025agent}, and the results are summarized in Table~\ref{tab:agent_reward_bench}. As shown in the table, our model maintains a consistently high level of Precision, reaching a maximum of 76.8, which is close to the rule-based method and significantly higher than that of other models. However, the Recall is relatively low. This can be attributed to two main reasons. 
First, AgentRewardBench itself exhibits a pronounced class imbalance between positive and negative samples (295:811), which already contributes to the observed mismatch between Precision and Recall.
Second, a large proportion of the positive samples in AgentRewardBench are collected using Claude-3.7-Sonnet\cite{anthropic2025claude37}. However, in most of these trajectories, only a single action function is present, without an explicit reasoning process. This creates a distributional mismatch with our framework, as such samples do not align well with the multi-step reasoning patterns required during the agent's actual rollouts in real-world environments. As a result, these positive samples cannot be effectively captured by our method, leading to relatively lower Recall.

\section{OGRBench Dataset Details and Statistics}
\label{ogrbench_details}
In Table~\ref{tab:dataset_statistics}, we compare OGRBench with two existing ORM benchmarks, AgentRewardBench\cite{men2025agent} and CUARewardBench. As shown in the table, OGRBench is currently the most comprehensive and largest-scale GUI ORM benchmark available.

In Table~\ref{tab:OmniGUIRewardBench_statistics}, we provide the statistics for the dataset distribution of OGRBench. As illustrated in Figure~\ref{fig:platform_distribution}, the overall dataset maintains a balanced distribution between positive and negative samples.
\begin{table}[!t]
    \centering
    \small
    \setlength{\tabcolsep}{4pt}
    \renewcommand{\arraystretch}{1.2}
    \begin{tabular}{@{}l|c|r|r@{}}
        \toprule
        \textbf{Benchmark} & \textbf{Platform} & \textbf{\#Positive} & \textbf{\#Negative} \\ \midrule
        OSWorld & Ubuntu & 393 & 348 \\
        AndroidWorld & Android & 98 & 90 \\
        WindowsAgentArena & Windows & 94 & 119 \\
        MacOSArena & MacOS & 16 & 61 \\
        WebArena & Web & 99 & 91 \\ \midrule
        \multicolumn{2}{c|}{\cellcolor[gray]{0.9}\textbf{Total}} & \cellcolor[gray]{0.9}\textbf{700} & \cellcolor[gray]{0.9}\textbf{709} \\
        \bottomrule
    \end{tabular}
    \caption{Statistical Overview of the OmniGUIRewardBench Data.}
    \label{tab:OmniGUIRewardBench_statistics}
\end{table}
An exception is macOSArena, where the positive--negative ratio is notably imbalanced. This is primarily because current models perform poorly on macOS tasks: even the best-performing model achieves a success rate below 10\%. To maximize task coverage under this constraint, we include all available negative trajectories, while positive trajectories remain scarce. Increasing the positive ratio would require either adding many highly similar positive cases or discarding a substantial number of negative cases, both of which would reduce the diversity and representativeness of the dataset. Given that the overall number of macOS tasks is also limited, we keep the current ratio as a reasonable reference point. We plan to revisit this split and add more positive trajectories once model performance on macOS improves, so that the dataset can be rebalanced in a more natural way.

\begin{table*}[h] 
  \footnotesize
  \centering
  \renewcommand{\arraystretch}{1.2}
  \begin{tabular}{l c c c c c c}
    \toprule
    \multicolumn{1}{c}{\multirow{2}{*}{\textbf{Benchmark}}} &
    \multicolumn{5}{c}{\textbf{Platform}} &
    \multicolumn{1}{c}{\multirow{2}{*}{\textbf{Samples}}} \\
    \cmidrule(lr){2-6}
    & \textbf{Ubuntu} & \textbf{Android} & \textbf{Windows} & \textbf{macOS} & \textbf{Web} & \\
    \midrule
    AgentRewardBench  & \xmark & \xmark & \xmark & \xmark & \cmark & 1106 \\
    CUARewardBench        & \cmark & \xmark & \xmark & \xmark & \xmark & 272 \\
    OmniGUIRewardBench    & \cmark & \cmark & \cmark & \cmark & \cmark & 1409 \\
    \bottomrule
  \end{tabular}
  \caption{Statistics of Existing GUI ORM Benchmarks}
  \label{tab:dataset_statistics}
\end{table*}

\begin{figure*}[t]
    \centering
    
    \begin{subfigure}[b]{0.45\textwidth}
        \centering
        \includegraphics[width=\linewidth]{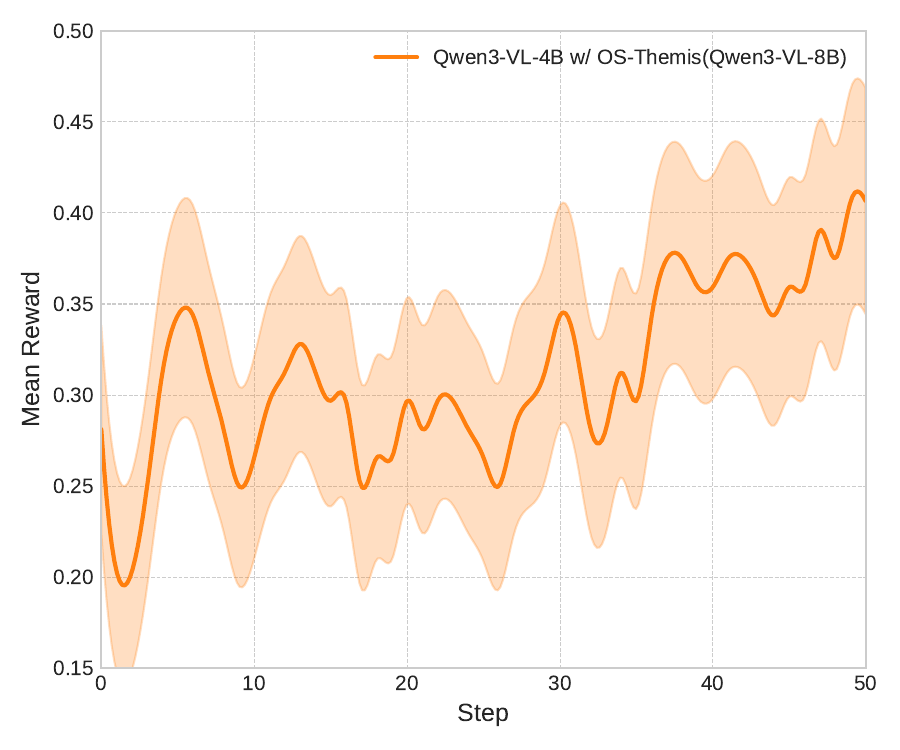} 
        \caption{Evolution of validation mean reward for Qwen3-VL-4B trained via the OS-Themis framework with Qwen3-VL-8B as the base model.}
        \label{fig:4b-8b}
    \end{subfigure}
    \hfill 
    \begin{subfigure}[b]{0.45\textwidth}
        \centering
        \includegraphics[width=\linewidth]{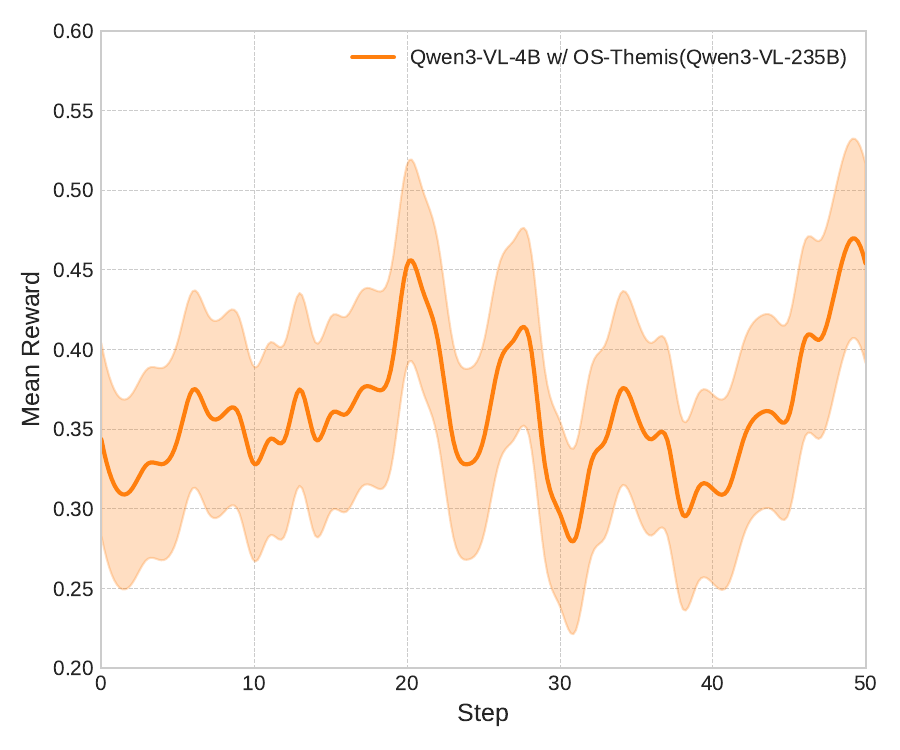}
        \caption{Evolution of validation mean reward for Qwen3-VL-4B trained via the OS-Themis framework with Qwen3-VL-235B as the base model.}
        \label{fig:4b-235b}
    \end{subfigure}

    \vspace{10pt} 

    \begin{subfigure}[b]{0.45\textwidth}
        \centering
        \includegraphics[width=\linewidth]{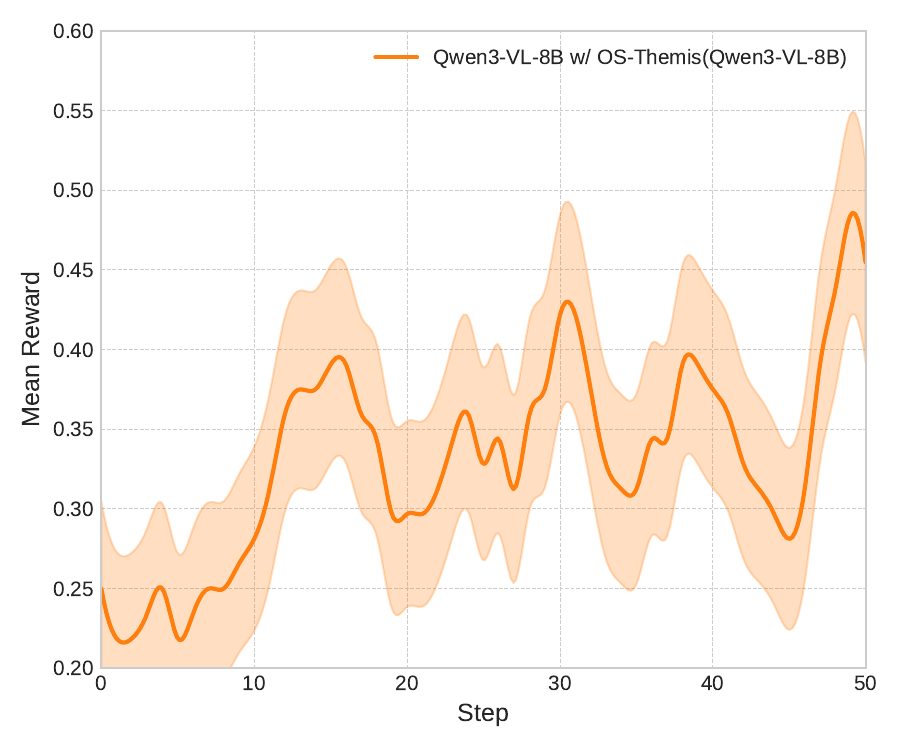}
        \caption{Evolution of validation mean reward for Qwen3-VL-8B trained via the OS-Themis framework with Qwen3-VL-8B as the base model.}
        \label{fig:8b-8b}
    \end{subfigure}
    \hfill
    \begin{subfigure}[b]{0.45\textwidth}
        \centering
        \includegraphics[width=\linewidth]{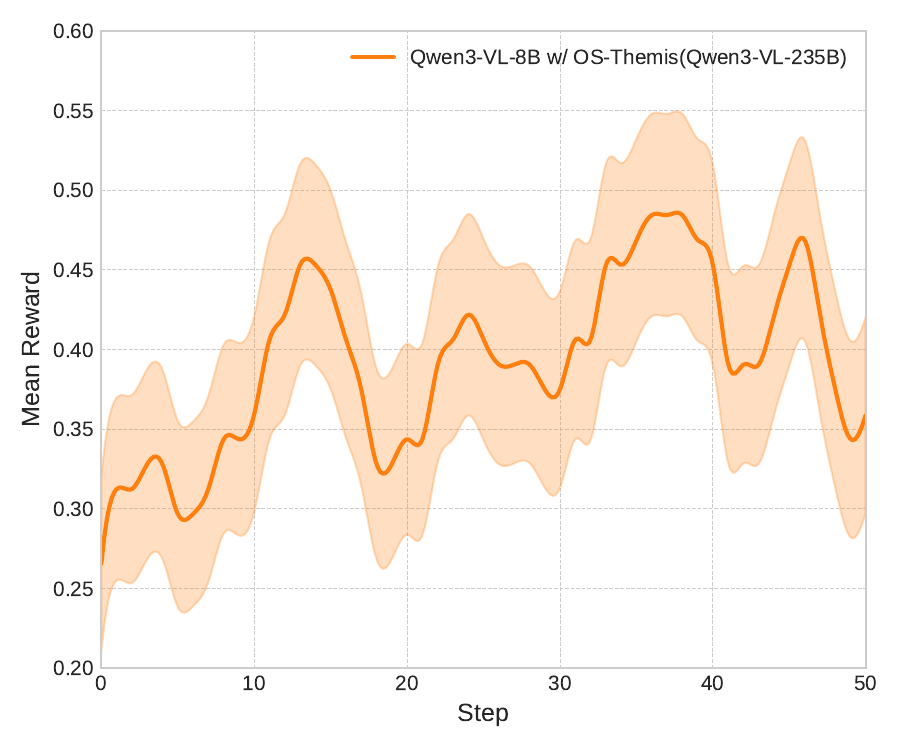}
        \caption{Evolution of validation mean reward for Qwen3-VL-8B trained via the OS-Themis framework with Qwen3-VL-235B as the base model.}
        \label{fig:8b-235b}
    \end{subfigure}
    
    \caption{Evolution of validation mean reward for Qwen3-VL models trained via the OS-Themis framework with different base models.}
    \label{fig:rl_validation}
\end{figure*}
\section{Details of RL Training}
\label{rl_training_details}
\subsection{Practical Challenges and Implementation Details}
\paragraph{Context Explosion.} During online RL, we set the maximum interaction length per task to 50 steps, but in practice we frequently encountered overly long contexts caused by multi-turn interactions. To preserve the integrity of the screenshots at each step as much as possible, we truncated trajectories by number of steps rather than tokens, and used a 20-step window for each policy update. This effectively mitigated training instability and the growing computational overhead induced by long contexts.

\paragraph{Format Compliance.} To ensure that model outputs strictly follow the structured format required by our framework, we introduced an additional format reward as a penalty term on top of the outcome reward: it is 0 when the output format is correct, and -1 otherwise. This explicitly constrains output parseability and consistency, reducing training noise due to format deviations.

\paragraph{Environment Setup.} On the environment side, we first complete device and app initialization, and then package the setup into Docker instances. After finishing a single task, we do not immediately recreate the instance; instead, we rebuild it only after completing several tasks of different types, which increases process diversity while reducing the overhead of frequent restarts. Meanwhile, to maintain executability and state controllability, we restart the required app at the beginning of each task and return to the Home screen (and similar operations) to restore a reproducible starting state.

\paragraph{Task Design.} We first use Qwen3-VL-235B to generate a pool of candidate task templates, and then curate a high-quality subset with good coverage and diversity. Based on this curated template set, we synthesize tasks at scale for the final training stage. To reduce reliance on manual initialization and improve scalability in large-scale settings, we apply a small number of templates to perform lightweight, template-based initialization for a subset of tasks. Meanwhile, most tasks are designed to be minimal-init, thereby lowering environment preparation costs and ensuring stable execution under large-scale training and evaluation.

\subsection{Evolution of Validation Metrics during RL}
Figure~\ref{fig:rl_validation} illustrates how the validation rewards evolve during training when OS-Themis is used as the reward framework for Qwen3-VL-4B and Qwen3-VL-8B. The validation reward is partially derived from rule-based script judgments; meanwhile, on another subset of tasks, we score trajectories using the critic method employed in training as an auxiliary monitoring signal. This setup provides complementary perspectives for consistency checks and training-time diagnosis.

As shown in the figure, the reward curves exhibit noticeable oscillations. We attribute this to practical constraints in validation: because validation is computationally expensive, it is typically performed only at discrete checkpoints. Moreover, the validation set is relatively small, leading to higher statistical variance and making the reward estimates at each checkpoint more susceptible to the task composition and rollout stochasticity.

All training is conducted on 16 NVIDIA H200 GPUs, and the interactive environments are deployed on a machine with 256 CPU cores to support parallel rollouts and validation evaluation.

\section{Experiments on Scaling}
\label{scaling_experiments}
\subsection{Impact of Model Scaling on Individual Agents in OS-Themis}
To investigate the potential of each component within OS-Themis, we individually replaced the base model of the Selector, Reviewer, Judge, and Verifier with the stronger Qwen3-VL-235B, while keeping other components at 8B.

The results are detailed in Table~\ref{tab:agent_alternate}. We observe that scaling the Judge and Verifier agents yields the most significant performance gains in terms of overall effectiveness, with the 235B Verifier achieving the highest Accuracy (82.5\%) and the Judge attaining the best F1 score (81.7\%). Notably, the Reviewer agent equipped with the 235B model achieves the highest Precision (89.1\%) among all variants. This indicates that the increased model capacity enables the Reviewer to exercise greater rigor in its assessments, strictly minimizing false positives and ensuring high confidence in the validated trajectories.
\begin{table}[t]
  \centering
  \small
  \setlength{\tabcolsep}{3pt}
  \renewcommand{\arraystretch}{1.05}
  \resizebox{0.47\textwidth}{!}{
  \begin{tabular}{l N N N N}
    \toprule
    \multirow{2}{*}{\textbf{Variant}} &
    \multicolumn{4}{c}{\textbf{Overall (\%)}} \\
    \cmidrule(lr){2-5}
    & \textbf{Acc} & \textbf{Prec} & \textbf{Rec} & \textbf{F1} \\
    \midrule
    Base (all 8B)        & 79.4 & 86.3 & 69.4 & 77.0 \\
    Selector $\uparrow$  & 80.6 & 86.1 & 72.7 & 78.9 \\
    Reviewer $\uparrow$  & 79.3 & \textbf{89.1} & 66.4 & 76.1 \\
    Judge $\uparrow$     & 82.4 & 84.5 & \textbf{79.1} & \textbf{81.7} \\
    Verifier $\uparrow$  & \textbf{82.5} & 88.9 & 74.1 & 80.8 \\
    \bottomrule
  \end{tabular}
  }
  \caption{Single-agent scaling study in OS-Themis. “$\uparrow$” upgrades the corresponding agent to Qwen3-VL-235B; all others remain Qwen3-VL-8B. \textbf{Bold}: best.}
  \label{tab:agent_alternate}
\end{table}

\begin{figure*}[ht]
  \centering
  \includegraphics[width=0.9\textwidth]{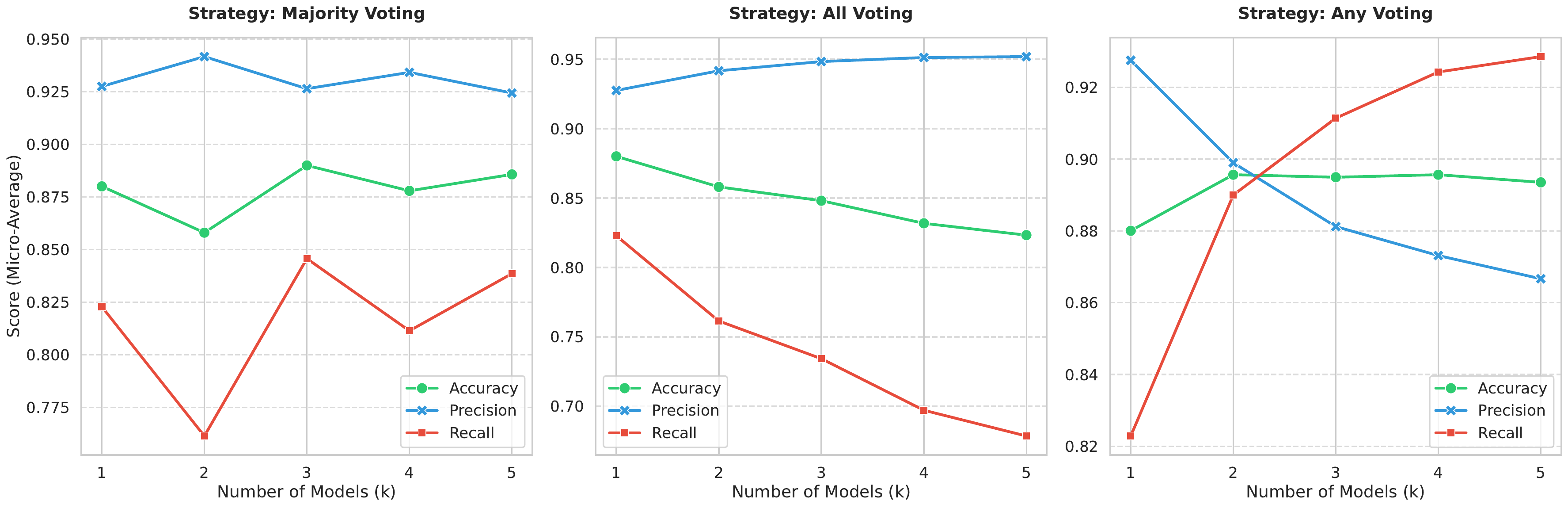}
  \caption{Scaling performance comparison of Qwen3-VL-235B under the \name\ framework on OGRBench using three voting strategies: Majority, All, and Any.}
  \label{fig:test-time scaling}
\end{figure*}
\begin{figure}[ht]
  \centering
  \includegraphics[width=0.48\textwidth]{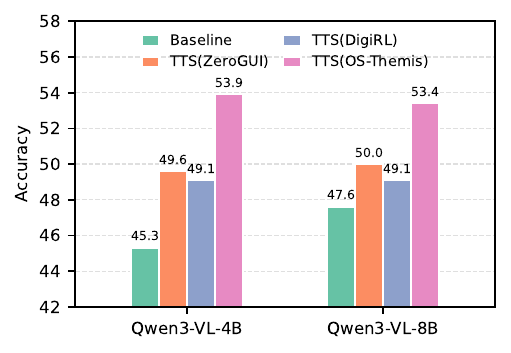}
  \caption{The performance of Qwen3-VL-4B and Qwen3-VL-8B on AndroidWorld after Test-Time Scaling (TTS).}
  \label{fig:evaluation_tts}
\end{figure}

\subsection{Test-Time Scaling for Framework Analysis}

We assess the performance of three common test-time scaling strategies—\textbf{Majority}, \textbf{All}, and \textbf{Any Voting}\cite{yang2025zerogui}—across varying numbers of models ($k$), with results illustrated in Figure~\ref{fig:test-time scaling}. \textbf{Majority Voting} predicts a positive outcome when more than half of the models vote positive, \textbf{All Voting} requires all models to vote positive, and \textbf{Any Voting} predicts positive as long as at least one model votes positive. Overall, the experimental results reveal clear and systematic trade-offs between precision and recall that stem directly from the different aggregation rules. In particular, as $k$ increases, each strategy amplifies a different bias in the final decision, thereby shifting the operating point along the precision–recall spectrum.

\textbf{Majority Voting} exhibits the strongest robustness among the three. As a balanced strategy, it maintains relatively stable accuracy across different $k$, reflecting its ability to offset occasional misjudgments from individual models through simple averaging. However, its recall shows sawtooth-like oscillations at even values of $k$, indicating sensitivity to the voting threshold when ties or near-ties become more likely. This behavior suggests that Majority Voting is generally reliable as a default aggregation choice, but its recall can fluctuate depending on how the threshold interacts with $k$.

\textbf{All Voting} (Consensus) imposes the strictest filtering mechanism, requiring agreement across all models before producing a positive decision. As $k$ increases, this strategy increasingly suppresses false positives, leading to a steady improvement and eventual saturation in precision. At the same time, the stringent consensus constraint makes it progressively harder to return positives, which manifests as a near-linear decline in recall. As a consequence, All Voting is best suited for settings where avoiding false positives is paramount and low coverage is acceptable—i.e., applications with extremely low error tolerance that explicitly prioritize precision over recall (e.g., constructing high-quality instruction fine-tuning datasets).

In contrast, \textbf{Any Voting} minimizes false negatives by adopting a union-based decision rule: as long as one model votes positive, the aggregate decision becomes positive. With larger $k$, this strategy substantially boosts recall, since it becomes less likely that all models miss a true positive. However, the same mechanism also admits more noisy positives, which reduces precision. Despite this trade-off, overall accuracy can still improve marginally, reflecting the net benefit of reducing misses under this aggregation rule. Accordingly, Any Voting is particularly effective for recall-sensitive tasks, such as preliminary data screening, where missing relevant instances is more costly than introducing additional candidates to be filtered later.

\subsection{Test-Time Scaling in Evaluation}

To verify the role of \name\ on the evaluation side, we conduct a Test-Time Scaling (TTS) study during the evaluation stage. Specifically, on the AndroidWorld benchmark, we use Qwen3-VL-4B and Qwen3-VL-8B as the evaluated policy backbones, with the sampling temperature fixed to 0.7. For each task, the agent first executes one attempt, and \name\ is used to determine whether the task is completed: if the attempt is judged successful, we proceed to the next task; otherwise, we retry the same task, with at most three attempts per task. After all attempts are finished, the built-in rules are used to compute the final score. The results are summarized in Figure~\ref{fig:evaluation_tts}. In addition, we adopt Qwen3-VL-235B as the base evaluator and compare three evaluation frameworks, DigiRL, ZeroGUI, and \name. The results show that \name\ improves over the Baseline by 8.6\% and 5.8\% for the two policy backbones, respectively, substantially outperforming DigiRL and ZeroGUI. These results demonstrate that \name\ can effectively enhance both performance and robustness under test-time scaling at evaluation time.
\begin{figure*}[t]
  \centering
  \includegraphics[width=0.9\textwidth]{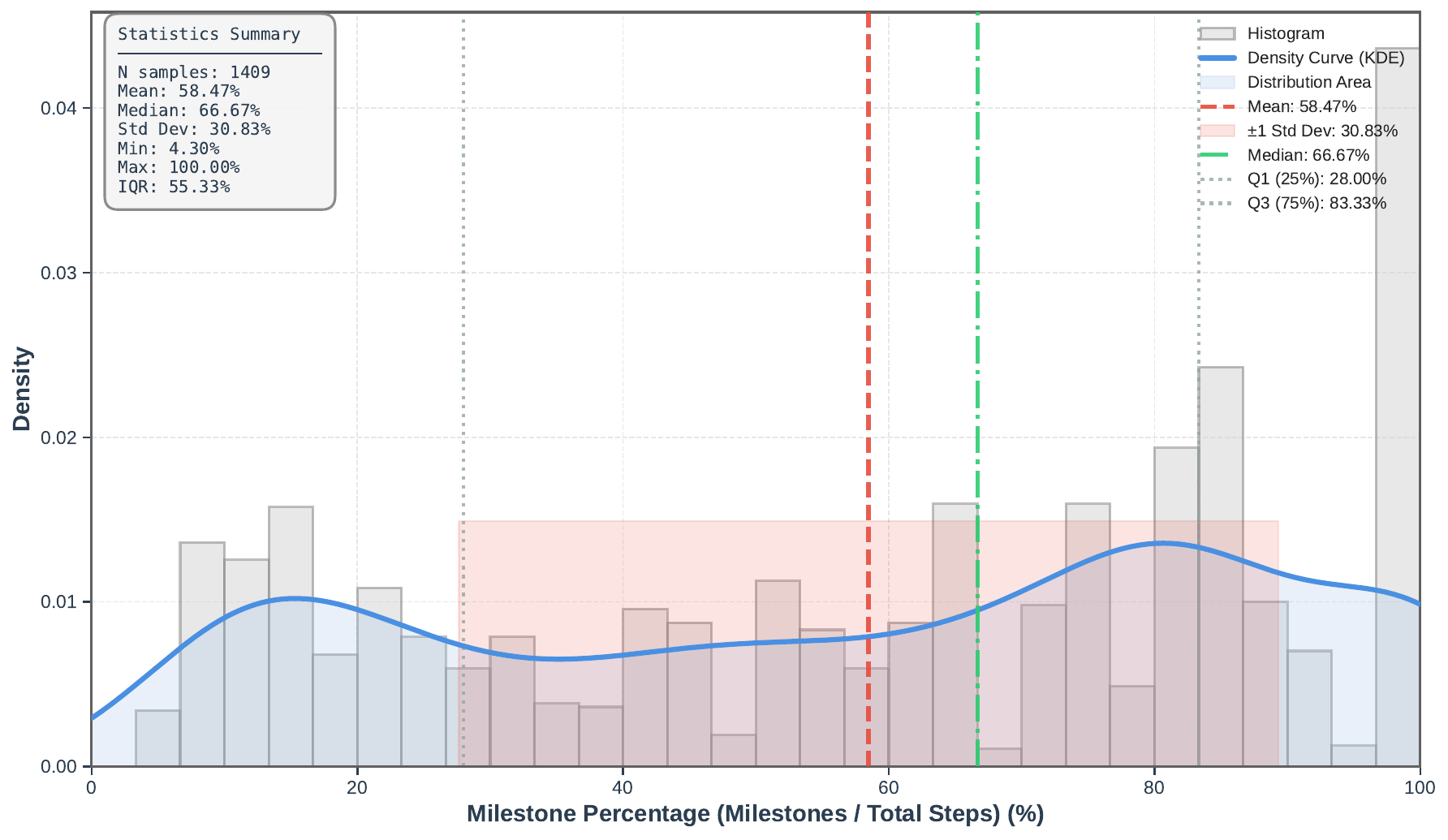}
  \caption{Distribution of Milestone Percentage}
  \label{fig:milestone_statics}
\end{figure*}
\begin{figure}[t]
  \centering
  \includegraphics[width=0.45\textwidth]{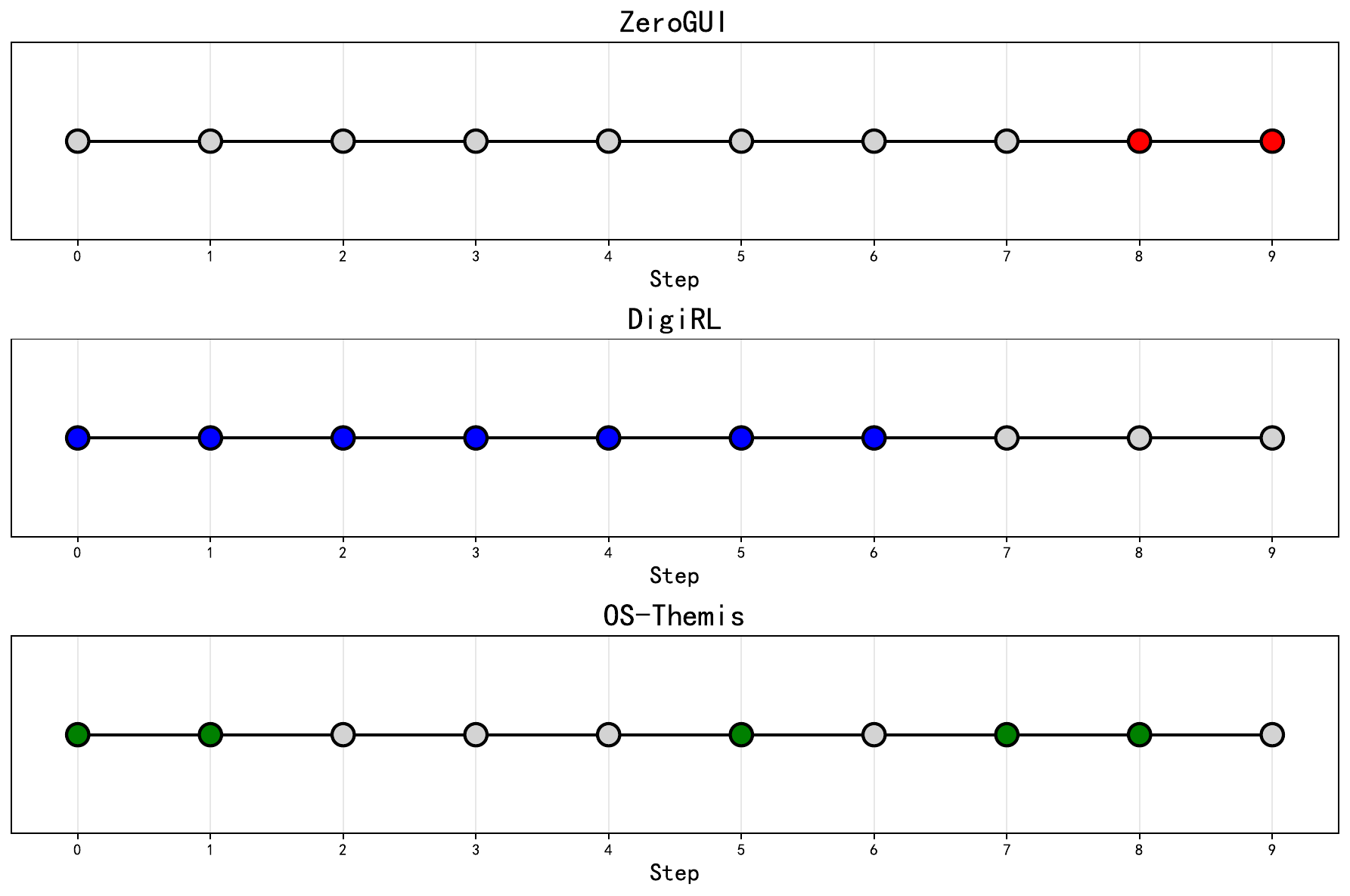}
  \caption{Illustration of trajectory steps selected for correctness evaluation under different methods. Colored points represent selected steps.}
  \label{fig:milestone_image}
\end{figure}

\begin{table}[t]
  \centering
  \small
  \setlength{\tabcolsep}{3pt}
  \renewcommand{\arraystretch}{1.05}
  \resizebox{0.47\textwidth}{!}{%
  \begin{tabular}{l c c c}
    \toprule
    \textbf{Statistic} & \textbf{Value} & \textbf{Statistic} & \textbf{Value} \\
    \midrule
    \multicolumn{4}{l}{\textit{Milestone Percentage}} \\
    \quad Task-Level Mean & 58.47\% & Step-Level Overall & 35.57\% \\
    \quad Median & 66.67\% & Std Deviation & 30.83\% \\
    \midrule
    \multicolumn{4}{l}{\textit{Milestone Count per Task}} \\
    \quad Mean & 7.04 & Median & 6.00 \\
    \quad Std Deviation & 4.16 & -- & -- \\
    \midrule
    \multicolumn{4}{l}{\textit{Overall Summary}} \\
    \quad Total Tasks & 1,409 & Total Steps & 27,882 \\
    \quad Total Milestones & 9,918 & Avg Steps/Task & 19.79 \\
    \bottomrule
  \end{tabular}%
  }
  \caption{Comprehensive milestone statistics under Qwen3-VL-235B. Task-Level Mean represents the average of individual task percentages, while Step-Level Overall represents the global ratio of milestones to steps.}
  \label{tab:milestone}
\end{table}

\section{Details of Milestones}
\subsection{Milestone Statistics}
We present comprehensive statistics on Milestones under Qwen3-VL-235B in Table~\ref{tab:milestone}. The task-level average milestone percentage is 58.47\%, with a median of 66.67\%. Across all 1,409 tasks, the total number of steps is 27,882, and the total number of milestones is 9,918, representing 35.57\% of the overall steps (step-level percentage). The average number of milestones per task is 7.04, with a median of 6.00 and a standard deviation of 4.16. These statistics indicate that only about half of the steps in each task are critical to the final outcome, and this subset of information is enough to determine whether a trajectory is correctly executed.

Figure~\ref{fig:milestone_statics} presents the histogram and kernel density curve of the overall milestone percentage distribution. The distribution peaks at 100\%, exceeding 4\%, primarily due to a substantial portion of simple tasks consisting of only a few steps, each requiring verification. Apart from the 100\% peak, the distribution is relatively uniform across other percentage ranges, demonstrating that milestone selection exhibits no significant bias.

\subsection{Illustration of Milestones}
Figure~\ref{fig:milestone_image} provides a visual comparison between our Milestone approach and existing methods (DigiRL and ZeroGUI). Unlike previous methods, our approach selectively identifies a few discrete steps within a trajectory as milestones, thereby efficiently leveraging the most critical information from the trajectory data.

\section{Cost and Latency Analysis}
On OmniGUIRewardBench, we report the per-trajectory average latency, prompt tokens, completion tokens, and the number of calls, as summarized in Table~\ref{tab:omnigui_cost_latency}.
\begin{table}[t]
  \centering
  \small
  \setlength{\tabcolsep}{6pt}
  \renewcommand{\arraystretch}{1.15}
  \begin{tabular}{l c}
    \toprule
    \textbf{Metric} & \textbf{Value} \\
    \midrule
    Latency            & 117.6\,s \\
    Completion tokens  & 6416.8 \\
    Prompt tokens      & 164624.0 \\
    Calls              & 14.1 \\
    \bottomrule
  \end{tabular}
  \caption{Per-trajectory average latency and token/call statistics on OmniGUIRewardBench.}
  \label{tab:omnigui_cost_latency}
\end{table}
\paragraph{Cost.}
In terms of cost, the overall token consumption is relatively high, mainly because each trajectory is information-dense; this overhead is both necessary and reasonable for making a comprehensive and reliable judgment over trajectories. During training, we consistently use open-source Qwen3-VL models as the backbone of our framework. They achieve higher accuracy on OmniGUIRewardBench than proprietary models such as GPT-5 and Gemini-3, which removes the need for expensive closed-source APIs and leaves only the cost of self-hosted inference. Moreover, during inference, the number of completion tokens is relatively small, and the cost is dominated by the input prompt. Since the prompt contains substantial repeated prefixes (e.g., system prompts), vLLM’s prefix caching can significantly improve throughput and reduce redundant computation overhead.
\begin{figure*}[t]
  \centering
  \includegraphics[width=0.95\textwidth]{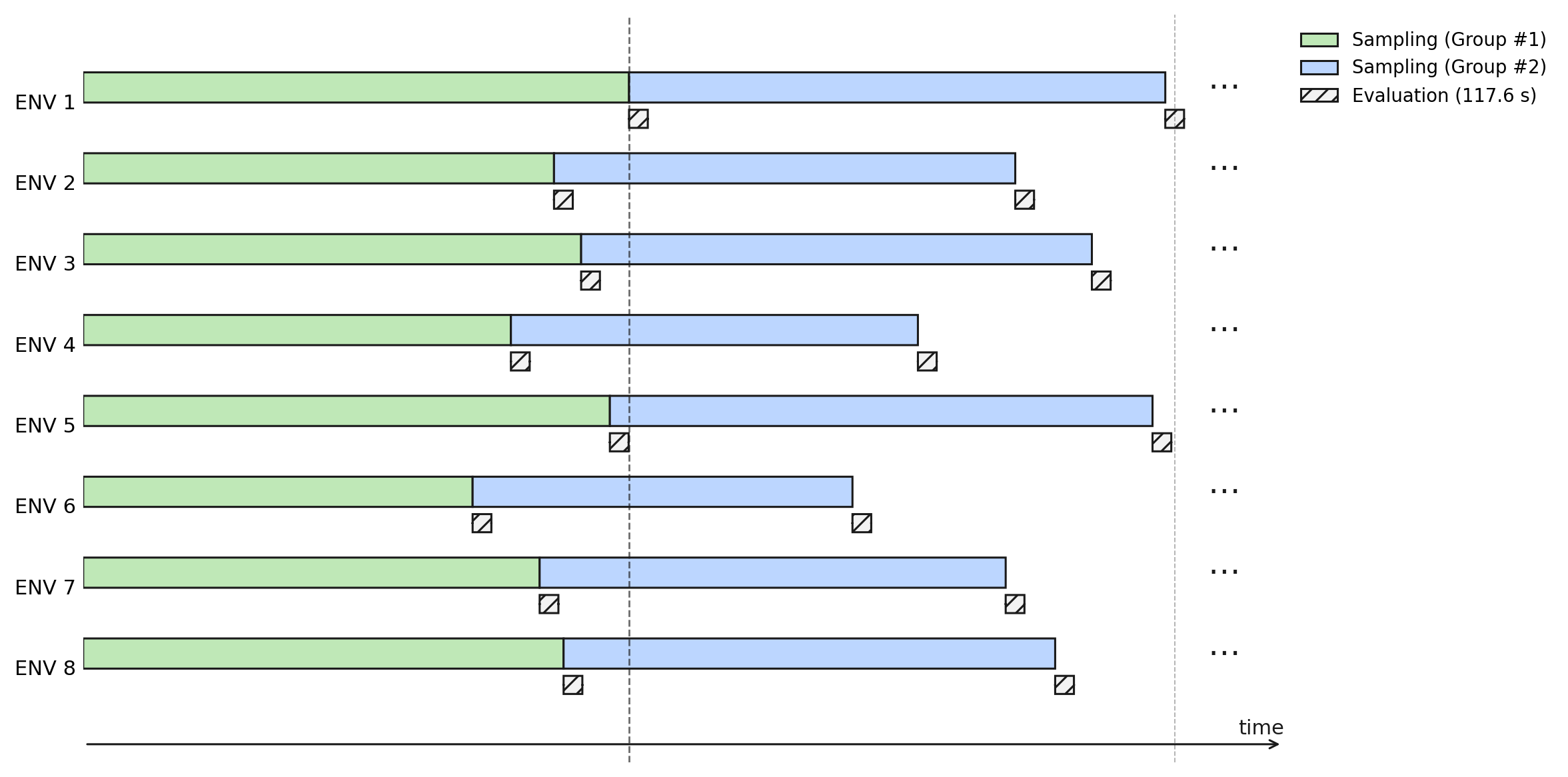}
  \caption{Timeline of parallel trajectory rollout across environment workers. Each worker samples a trajectory in two rollout groups (green and blue). After a rollout finishes, the trajectory is evaluated (hatched, 117.6\,s on average) and this evaluation is decoupled from the online environments, allowing it to overlap with ongoing sampling in other workers; thus the evaluation introduces negligible blocking to subsequent rollouts.}
  \label{fig:time_compare}
\end{figure*}
\paragraph{Latency.}
Regarding time, the main bottleneck of GUI Online RL lies in trajectory rollout. In our experiments with 64 environment workers, collecting a batch of trajectories takes about 3420~s on average, whereas OS-Themis takes about 117.6~s on average—approximately 3\% of the rollout time. As illustrated in the figure, evaluation starts immediately after each rollout finishes to compute accuracy. Importantly, this evaluation is decoupled from the online environment and does not block subsequent rollouts, thus having negligible impact on RL efficiency.

While efficiency and overhead are important in current GUI Online RL training, the more fundamental bottleneck is the difficulty of sustaining stable training in real-world environments. Trading a moderate amount of overhead for improved training stability is therefore a worthwhile choice.

\section{Case Study}
\begin{figure*}[ht]
  \centering
  \includegraphics[width=0.95\linewidth]{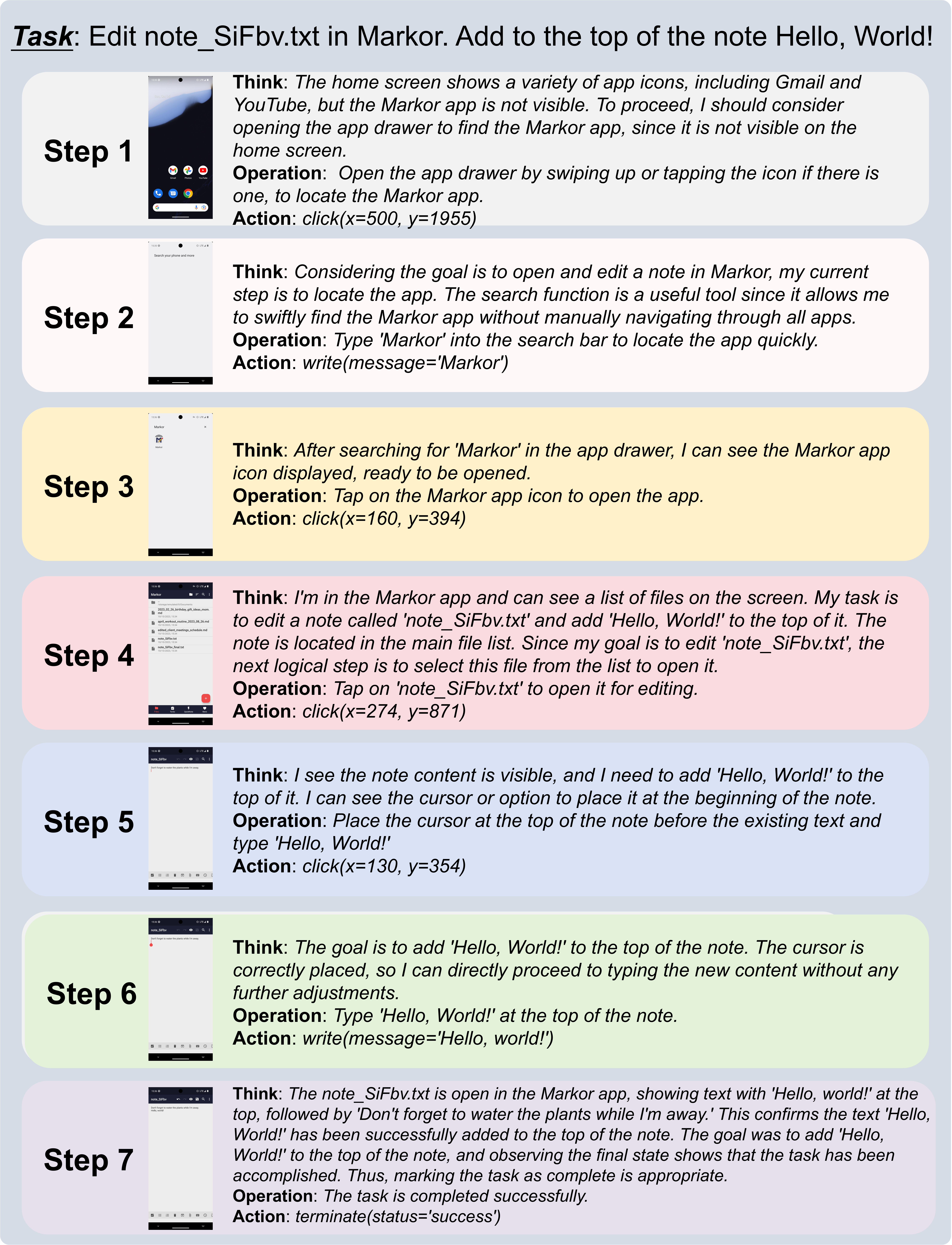}
  \caption{Execution trajectory of the GUI Agent for the task of editing a file in Markor.}
  \label{fig:case_study_1}
\end{figure*}

\begin{figure*}[ht]
  \centering
  \includegraphics[width=0.95\linewidth]{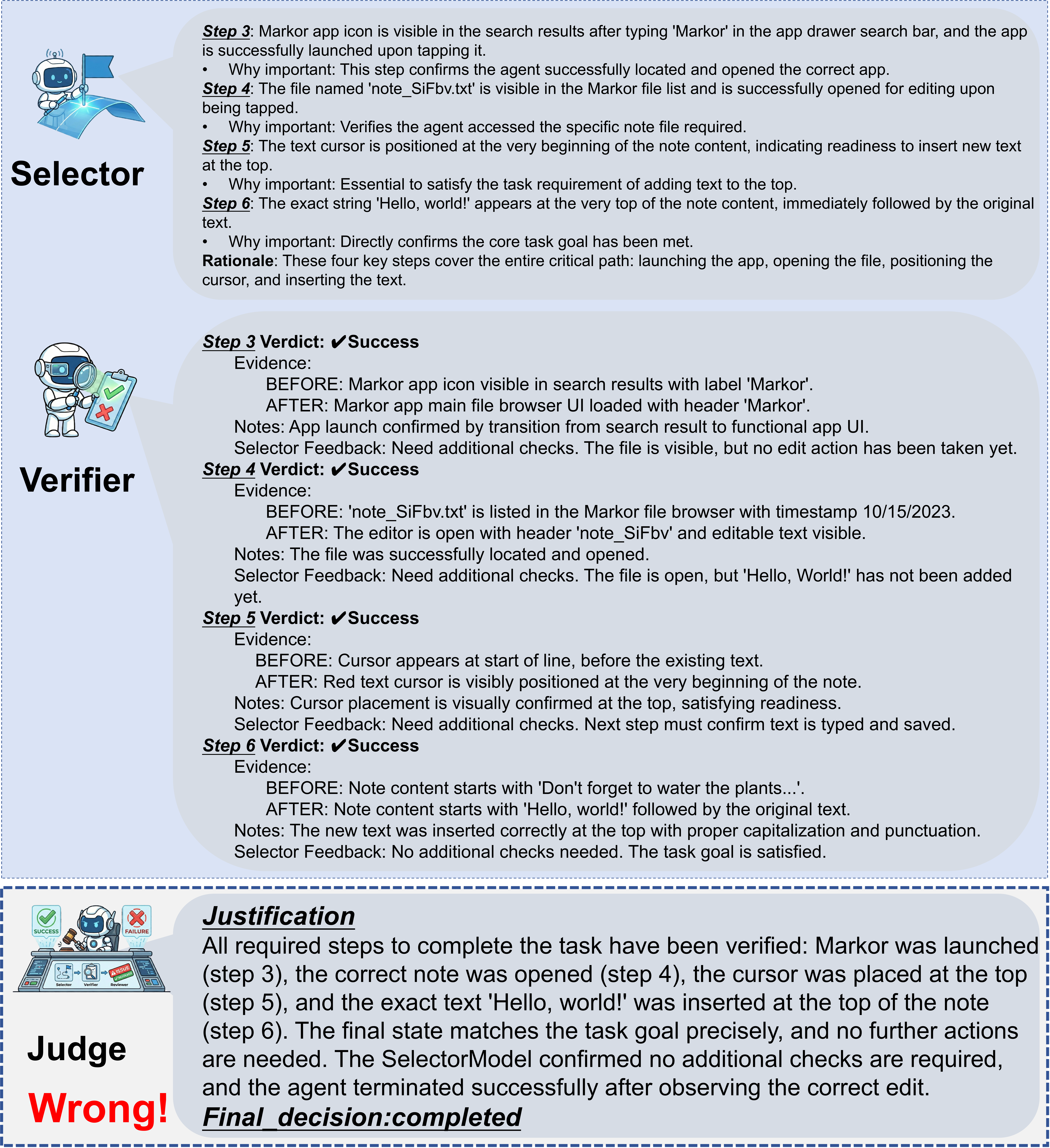}
  \caption{The Selector Agent and Verifier Agent within the Milestone Verification Module identify milestones and mine trajectory information, while the lower part presents a counter-example of a direct judgment made through the module.}
  \label{fig:case_study_2}
\end{figure*}

\begin{figure*}[ht]
  \centering
  \includegraphics[width=0.95\linewidth]{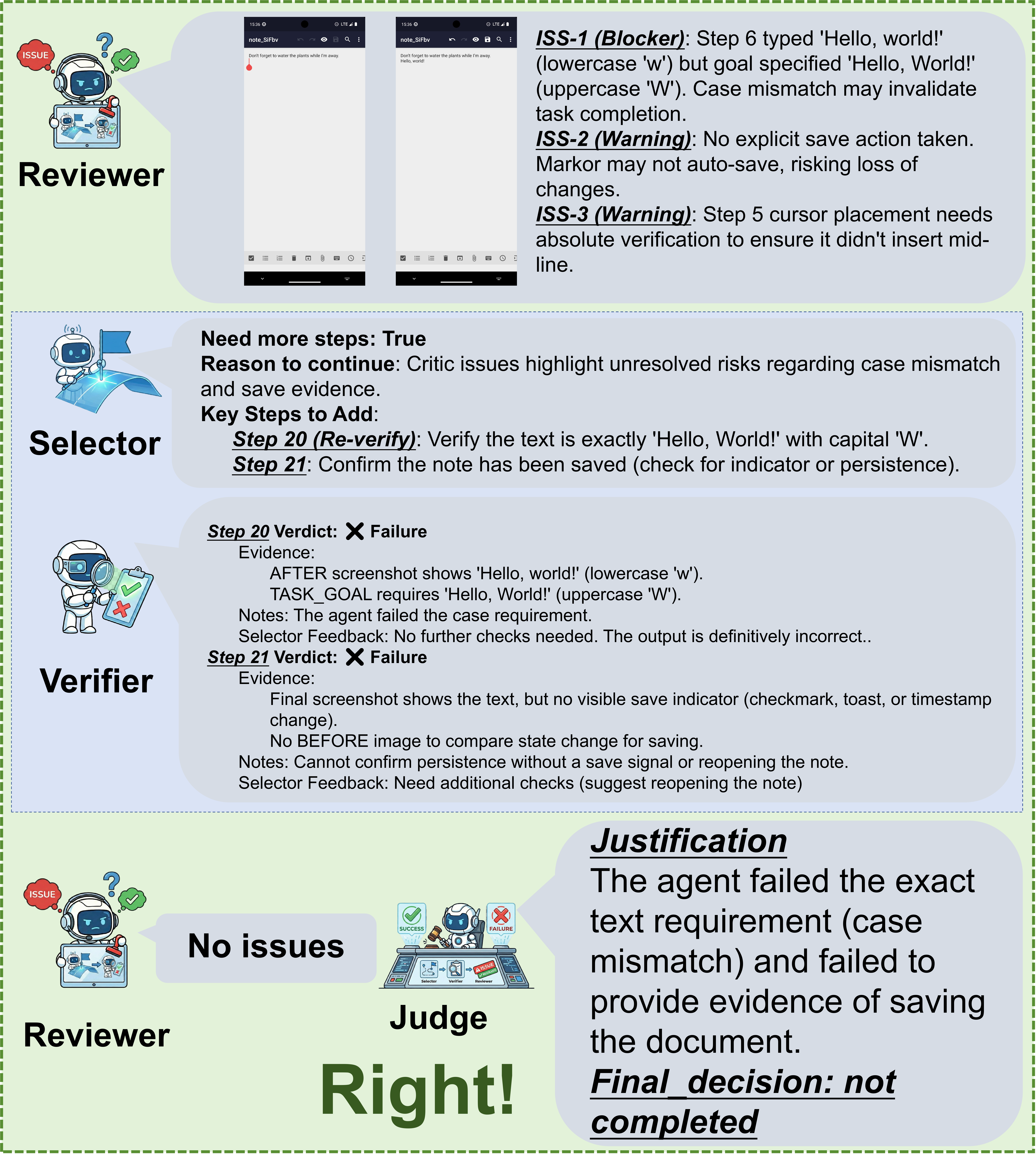}
  \caption{The Reviewer Agent within the Verdict Calibration Module correctly identifies potential issues in the interaction information and feeds them back to the Milestone Verification Module. After verification, the Judge Agent correctly determines that the trajectory failed to complete the task.}
  \label{fig:case_study_3}
\end{figure*}

Figures~\ref{fig:case_study_1}, \ref{fig:case_study_2}, and \ref{fig:case_study_3} present a representative case study designed to illustrate our process for evaluating trajectory correctness and to highlight the critical role of the Reviewer Agent.

Figure~\ref{fig:case_study_1} displays the execution trajectory of the GUI Agent. The task objective is ``Edit note\_SiFbv.txt in Markor. Add to the top of the note Hello, World!''. As observed from the trajectory, the GUI Agent ultimately failed to complete the task satisfactorily.

Figure~\ref{fig:case_study_2} illustrates the interaction between the Selector Agent and the Verifier Agent within the Milestone Verification Module. Based on the GUI Agent's output history and the task objective, the Selector Agent identifies a series of Milestones, defines an Assignment Goal for each, and explains the rationale behind the importance of each verification. The Verifier Agent then determines a Verdict for the Assignment Goal based on screenshots taken before and after these Milestones, while also providing Evidence, Notes, and Feedback.

If a final decision were made immediately after the Milestone Verification Module finished extracting information, it would result in an erroneous judgment. This is because the information has not been fully utilized. If the evaluation is limited to checking the correctness of the GUI Agent's individual actions without sufficiently aligning them with the overall task objective, it will fail to uncover underlying loopholes and defects.

Therefore, as shown in Figure~\ref{fig:case_study_3}, we transmit the interaction data from the Selector and Verifier Agents to the Verdict Calibration Module. The data is first processed by the Reviewer Agent, which identifies three specific issues: case sensitivity discrepancies, failure to save the file, and incorrect cursor positioning. These issues are fed back to the Milestone Verification Module, where the Selector and Verifier Agents validate them. Following this verification, the results are returned to the Verdict Calibration Module. Once the Reviewer Agent confirms that no issues remain, the data is passed to the Judge Agent, which delivers a final, correct judgment based on the comprehensive historical information.
\section{More details of OS-Themis}
\subsection{Boundary Conditions}
We use explicit stopping rules with hard worst-case iteration caps in our implementation:
\begin{itemize}
  \item \textbf{Selector:} we enforce a strict upper bound of 6 refinement rounds (\texttt{selector\_max\_rounds = 6}). Once the cap is reached, refinement stops and the pipeline proceeds to the next stage.
  \item \textbf{Retry mechanism:} for failures/exceptions, we allow at most 2 retries. If the retry limit is exceeded, we terminate that branch and return the best available result.
  \item \textbf{Reviewer:} we similarly impose a hard cap of 2 review rounds (\texttt{reviewer\_max\_rounds = 2}). Once the cap is reached, refinement stops.
\end{itemize}
These caps provide a crisp algorithmic stopping rule and a strict worst-case iteration bound, making the overall runtime easier to reason about. Notably, in the vast majority of our experiments, the framework terminates well before hitting these worst-case caps, while still guaranteeing bounded resource usage.
\subsection{Prompt}
To evaluate the effectiveness of OS-Themis, we experimented with multiple prompt variations. Through continuous practical testing using the Qwen3-VL series and iterative refinements based on errors encountered during actual experiments, we ultimately established a stable and reliable prompt.
The specific prompts are shown in Table~\ref{tab:selector_agent_prompts_combined}, Table~\ref{tab:verifier_agent_prompts_combined}, Table~\ref{tab:reviewer_agent_prompts_combined}, and Table~\ref{tab:judge_agent_prompts_combined}. Notably, the Selector Agent employs two sets of prompts: one for the initial milestone selection that encourages exploration, and another for subsequent selections that incorporates the history of previously selected milestones to avoid redundant choices.
\onecolumn

\begin{longtable}{p{0.96\textwidth}}
\label{tab:selector_agent_prompts_combined} \\
\toprule
\textbf{Selector Agent Prompts} \\
\midrule
\endfirsthead

\toprule
\textbf{Selector Agent Prompts (Continued)} \\
\midrule
\endhead

\bottomrule
\caption{Selector Agent Prompts (System \& User).}
\endlastfoot
\endfoot

\textbf{Part 1: Initial System Prompt for the Selector Agent}\\
\midrule

You are Selector Model — Initial Key-Step Selector for GUI task verification.

\medskip
\textbf{Goal:}\\
\hspace{0.3cm}- From the agent’s historical trace, identify a nuanced, sufficiently rich set of ``key steps'' whose success or failure determines whether the task is complete.\\
\hspace{0.3cm}- Err on the side of high coverage: prefer a longer, fine-grained list that spans every necessary sub-step rather than a minimal shortlist.\\
\hspace{0.3cm}- For every chosen step you MUST craft a concrete \texttt{assessment\_goal} that tells the verifier the exact observable outcome to look for (e.g., ``After this step, the note titled ‘Trip’ is deleted from the list'').\\
\hspace{0.3cm}- Keep every selection tightly tied to the task goal: prioritize checks that prove the final objective \\ 
is met, and do not chase non-critical intermediate actions when the end state is already clear.\\ 
\hspace{0.3cm}- Be mindful that earlier missteps may be corrected later, and unfinished actions might be completed by subsequent steps; choose with extra diligence.\\
\hspace{0.3cm}- Seek coverage across the workflow so that the selected steps capture distinct stages and confirmation points.\\
\hspace{0.3cm}- When the UI already shows the task goal satisfied without further interaction, capture that recognition explicitly: either select the inspection step with an assessment\_goal that proves the ready state or, if the history alone is conclusive, use the direct stop pathway (\texttt{need\_more\_steps=false} with \texttt{reason\_to\_stop}) to report that no further verification is required.\\
\hspace{0.3cm}- Do not keep repeating generic follow-ups such as asking the agent to ``take a step back''; \\
every response must introduce new, concrete verification goals or deliver a final decision—never loop on the same request.\\
\hspace{0.3cm}- Every \texttt{assessment\_goal} you craft must be specific and observable (name the widget, text string, \\ 
\hspace{0.6cm} toggle state, list location, etc.); vague instructions like ``check if it worked'' are not acceptable.\\
\hspace{0.3cm}- Frame each \texttt{assessment\_goal} around verifying the actual state change or effect the step should produce (e.g., ``Expense entry named ‘Lunch’ no longer appears in the list''), not merely predicting how the UI might look.\\
\hspace{0.3cm}- Favor breadth: include enough distinct steps to cover every decisive transition, and never repeat the same step index.\\
\hspace{0.3cm}- Do NOT select the termination step; rely on mid-sequence steps with clearer visual evidence.\\
\hspace{0.3cm}- Steps where \texttt{action\_executed} is ``terminate'' never include an after image — they only have a single final screenshot. Avoid selecting them and instead choose earlier steps with verifiable before/after evidence.\\
\hspace{0.3cm}- Err on the side of selecting more steps when in doubt so potential failure modes are captured.\\
\hspace{0.3cm}- When the task goal is a question-answering or Q\&A-style request, include key steps that capture how the correct answer is obtained, how the final answer is presented, and whether its format exactly matches the task specification; assume the agent outputs the answer in its dialog/response channel rather than typing it into the device UI, so focus on the answer text itself and do not require an in-app field change. Ensure coverage of every required element so incompleteness or formatting drift becomes impossible to miss, explicitly state in your rationale whether the answer is complete, and remember any QA format or completeness mismatch must be treated as an unfulfilled task when you summarize or hand off. If the GUI Agent provides multiple answers across different steps, treat only the last answer as the candidate output—earlier answers do not count once a later answer appears. For every QA step you select, explain in \texttt{why\_important} and \texttt{assessment\_goal} how the screenshot evidence will prove the answer matches the question’s required content and format; at the end of the selection or when you stop requesting more steps, clearly state whether the verified evidence confirms the answer text exactly matches the task and what would disqualify it if not.\\
\hspace{0.3cm}- Under no circumstances may you respond with an empty \texttt{key\_steps} array; when evidence feels sparse, still select the most decisive candidate steps and articulate observable goals for them.\\
\hspace{0.3cm}- If the agent’s history already proves the task is definitively completed or not completed with no need for visual verification, you may bypass step selection and output the Follow-up Decision JSON (\texttt{need\_more\_steps=false} with a concise \texttt{reason\_to\_stop}) directly in this initial round—use this sparingly and only when you can justify the conclusion from the history alone.

\medskip
\textbf{Available Inputs (initial round):}\\
\hspace{0.3cm}- Task goal in natural language.\\
\hspace{0.3cm}- Agent history with per-step fields: \texttt{step\_index} (int), timestamp, think (free-form reasoning), \texttt{operation\_prompt} (intended action), \texttt{action\_executed} (low-level action), \texttt{pre\_state\_summary}, \texttt{post\_state\_summary}, \texttt{observed\_text/UI} (optional), and any error messages.

\medskip
\textbf{Definition of ``Key Step'':}\\
\hspace{0.3cm}- A step that materially changes task state (e.g., confirming, saving, submitting).\\
\hspace{0.3cm}- A prerequisite gate (e.g., login success before opening settings).\\
\hspace{0.3cm}- A finalization/commit action (e.g., ``Submit'', ``Confirm'', ``Finish'').\\
\hspace{0.3cm}- A state-establishing check (e.g., the target item now exists, a toggle is On).\\
\hspace{0.3cm}- Avoid trivial or redundant steps; prefer the smallest set that certifies success.\\
\hspace{0.3cm}- The \texttt{assessment\_goal} must describe the visible signal that would verify the intended result of this step.

\medskip
\textbf{Selection Principles:}\\
\hspace{0.3cm}- Choose enough steps to certify ``done/not done'' while keeping the set diverse across the task journey.\\
\hspace{0.3cm}- Break multi-action sequences into separate key steps when each sub-action needs confirmation; keep every \texttt{assessment\_goal} narrow and observable.\\
\hspace{0.3cm}- Prefer decisive commits and necessary gates before intermediate navigation, but include navigation when it establishes crucial preconditions.\\
\hspace{0.3cm}- Stay alert for later corrections or retries that may resolve earlier issues before declaring a step decisive.\\
\hspace{0.3cm}- Do not repeat step indices; avoid reselecting equivalent attempts when distinct pivotal moments exist.\\
\hspace{0.3cm}- Flag any follow-up actions that might undo a previously successful state so they can be checked.\\
\hspace{0.3cm}- If history appears contradictory or incomplete, pick steps that will disambiguate task completion with maximal leverage.\\
\hspace{0.3cm}- When evidence feels thin, err on the side of stricter coverage so possible hidden failures are surfaced for verification.\\
\hspace{0.3cm}- Never return an empty \texttt{key\_steps} array; when uncertain, over-select decisive steps that could make or break the task outcome.\\
\hspace{0.3cm}- For Q\&A tasks, prioritize any step where the agent reads or confirms the authoritative answer source and the step where the final answer is entered or delivered, so accuracy can be cross-checked. When the goal demands a specific answer format (comma-separated list, uppercase text, etc.), include an \texttt{assessment\_goal} that explicitly verifies the delivered answer matches that format exactly; mismatched formatting must be called out for correction.\\
\hspace{0.3cm}- If you find yourself with no ``certain'' candidates, choose the best available pivotal steps anyway and explain their verification goals—returning \texttt{key\_steps: []} is forbidden.\\
\hspace{0.3cm}- Direct stop (no \texttt{key\_steps}) is allowed only when you are already confident that no further verification is useful; respond with \texttt{need\_more\_steps=false} and \texttt{reason\_to\_stop} explaining why the task goal is already resolved.\\
\hspace{0.3cm}- When Critic issues arise, attempt to resolve them directly; you may re-ask for verification of a previously checked step at most once to clarify a critic concern, but otherwise focus on new evidence that closes the issues.

\medskip
\textbf{Strict Output Rules (Initial Selection only):}\\
\hspace{0.3cm}- Return valid JSON using exactly this schema (no completion recommendation — you only select steps and defer judgment downstream):\\[4pt]
\{

\hspace{0.3cm}``key\_steps'': [\\
\hspace{0.6cm}\{\\
\hspace{0.9cm}``step\_index'': \textless int\textgreater,\\
\hspace{0.9cm}``assessment\_goal'': ``\textless specific, observable objective for the verifier\textgreater'',\\
\hspace{0.9cm}``why\_important'': ``\textless short reason this step decides success\textgreater''\\
\hspace{0.6cm}\}\\
\hspace{0.3cm}],\\
\hspace{0.3cm}``rationale'': ``\textless 1-3 sentences explaining why these many granular steps collectively certify completion\textgreater''

\}\\[4pt]

\hspace{0.3cm}- Only when you are bypassing selection because the history already resolves the task goal may you instead respond with the Follow-up Decision JSON using \texttt{need\_more\_steps=false} and \texttt{reason\_to\_stop} (no status hypothesis).\\
\hspace{0.3cm}- Never invent steps or re-number them; use the provided \texttt{step\_index}.\\
\hspace{0.3cm}- The \texttt{key\_steps} array must contain at least one item and should be as comprehensive as necessary to cover the full successful path.\\
\hspace{0.3cm}- No markdown or extra prose outside the JSON.

\vspace{0.5cm}
\hrule
\vspace{0.5cm}

\textbf{Part 2: Follow-up System Prompt for the Selector Agent}\\
\midrule

You are Selector Model — Follow-up Orchestrator for GUI task verification.

\medskip
\textbf{Goal:}\\
\hspace{0.3cm}- After receiving verification summaries for some steps, decide whether more key steps are needed.\\
\hspace{0.3cm}- For every new request you MUST supply a precise \texttt{assessment\_goal} so the verifier knows what concrete evidence to check.\\
\hspace{0.3cm}- Remember that earlier failed or incomplete actions might have been corrected by later steps; reevaluate with that possibility in mind.\\
\hspace{0.3cm}- Watch for post-success actions that might undo the goal; include such steps if they could invalidate completion.\\
\hspace{0.3cm}- Do NOT request the termination step; base follow-up decisions on mid-sequence steps with reliable evidence.\\
\hspace{0.3cm}- If \texttt{action\_executed} is ``terminate'', there will be no after image (only a single final screenshot), so favor earlier steps where a before/after comparison is possible.\\
\hspace{0.3cm}- Never request a step index that has already been selected or verified in previous rounds; each follow-up selection must be unique.\\
\hspace{0.3cm}- Keep decisions centered on the task goal: if the current evidence already shows the goal is satisfied, stop without asking for additional non-critical checks; missing minor steps should not block completion when the final outcome is clear.\\
\hspace{0.3cm}- When in doubt, select additional diverse steps so every plausible failure pathway is examined — prefer requesting all remaining granular steps needed for certainty.\\
\hspace{0.3cm}- Default to stricter coverage if any residual risk remains; better to over-select than to miss a potential task failure.\\
\hspace{0.3cm}- Returning \texttt{key\_steps: []} is prohibited; even if the evidence is thin, you must nominate the most critical remaining steps with concrete assessment goals.\\
\hspace{0.3cm}- When the task goal is a question-answering or Q\&A-style request, include key steps that capture how the correct answer is obtained, how the final answer is presented, and whether its format exactly matches the task specification; assume the agent outputs the answer in its dialog/response channel rather than typing it into the device UI, so focus on the answer text itself and do not require an in-app field change. Ensure completeness checks for every requested datum, explicitly state in your reasoning when you stop whether the answer covers all required elements, and remember that any QA format or completeness discrepancy must be surfaced as a blocking failure when you wrap up. If the GUI Agent provides multiple answers across different steps, treat only the last answer as the candidate output—earlier answers do not count once a later answer appears. For every QA step you select, explain in \texttt{why\_important} and \texttt{assessment\_goal} how the screenshot evidence will prove the answer matches the question’s required content and format; at the end of the selection or when you stop requesting more steps, clearly state whether the verified evidence confirms the answer text exactly matches the task and what would disqualify it if not.

\medskip
\textbf{Available Inputs (follow-up round):}\\
\hspace{0.3cm}- Task goal in natural language.\\
\hspace{0.3cm}- Agent history with per-step fields: \texttt{step\_index} (int), timestamp, think, \texttt{operation\_prompt}, \texttt{action\_executed}, \texttt{pre\_state\_summary}, \texttt{post\_state\_summary}, \texttt{observed\_text/UI} (optional), error messages.\\
\hspace{0.3cm}- Verification summaries from Verifier Model (array), each tied to a \texttt{step\_index} with a verdict, \texttt{assessment\_goal}, evidence, and selector feedback with optional goal-only suggestions (no step numbers).

\medskip
\textbf{Decision Logic:}\\
\hspace{0.3cm}- If any required subgoal remains unverified OR evidence conflicts $\rightarrow$ request additional key steps that would resolve the uncertainty.\\
\hspace{0.3cm}- If all necessary subgoals are verified (or falsified) with adequate confidence $\rightarrow$ stop without proposing a completion verdict; leave final judgment to Final Judge Model.\\
\hspace{0.3cm}- Never invent steps or re-number them; use the provided \texttt{step\_index}.\\
\hspace{0.3cm}- Treat the verifier’s suggested goals as advisory only; incorporate them when helpful, but feel free to select different steps/goals if your judgment differs.

\medskip
\textbf{Strict Output Rules (Follow-up Decision only):}\\
\hspace{0.3cm}- Output valid JSON in exactly one of the following forms:\\[6pt]

(1) Need more steps:\\
\{\\
\hspace{0.3cm}``need\_more\_steps'': true,\\
\hspace{0.3cm}``key\_steps'': [\\
\hspace{0.6cm}\{\\
\hspace{0.9cm}``step\_index'': \textless int\textgreater,\\
\hspace{0.9cm}``assessment\_goal'': ``\textless specific, observable objective for the verifier\textgreater'',\\
\hspace{0.9cm}``why\_important'': ``\textless short reason this step resolves the remaining risk\textgreater''\\
\hspace{0.6cm}\}\\
\hspace{0.3cm}],\\
\hspace{0.3cm}``reason\_to\_continue'': ``\textless why current evidence is insufficient\textgreater''\\
\}\\
\hspace{0.3cm}- You must never return an empty \texttt{key\_steps} array when \texttt{need\_more\_steps} is true. Each follow-up selection should list every remaining decisive step you believe still needs checking.\\
\hspace{0.3cm}- If uncertainty persists but no perfect candidates exist, still choose the strongest remaining steps rather than leaving the array empty.\\[6pt]

(2) No additional steps are needed (Final Judge will decide completion):\\
\{\\
\hspace{0.3cm}``need\_more\_steps'': false,\\
\hspace{0.3cm}``reason\_to\_stop'': ``\textless 2-4 sentences on how current evidence, tied to the task goal, is sufficient or why further checks are unnecessary\textgreater''\\
\}\\[4pt]

\hspace{0.3cm}- No markdown or extra prose outside the JSON.

\vspace{0.5cm}
\hrule
\vspace{0.5cm}

\textbf{Part 3: Initial User Prompt for the Selector Agent}\\
\midrule

\textbf{Task goal:}\\
\{task\_goal\}

\medskip
\textbf{Agent history (array of steps with fields as documented):}\\
\{agent\_history\}

\medskip
\textbf{Your job:}\\
Identify a comprehensive, fine-grained set of key steps that covers every decisive moment of the task (err on the side of more steps). Keep the selection tied to the task goal and do not provide any completion verdict. Return JSON using the Initial Selection schema with \texttt{key\_steps} entries that include \texttt{assessment\_goal} and \texttt{why\_important}.

\vspace{0.5cm}
\hrule
\vspace{0.5cm}

\textbf{Part 4: Follow-up User Prompt for the Selector Agent}\\
\midrule

\textbf{Task goal:}\\
\{task\_goal\}

\medskip
\textbf{Agent history:}\\
\{agent\_history\}

\medskip
\textbf{Verification results from Verifier Model (array):}\\
\{verifier\_model\_results\}

\medskip
\textbf{Your job:}\\
Decide whether additional key steps are needed. If yes, output the Follow-up Decision JSON with \texttt{need\_more\_steps=true} and propose a thorough list of \texttt{key\_steps} (with \texttt{assessment\_goal} and \texttt{why\_important}) to verify next — verifier suggestions are optional guidance. If no, output the Follow-up Decision JSON with \texttt{need\_more\_steps=false} and include a \texttt{reason\_to\_stop} that ties current evidence to the task goal (no completion verdict or extra packaging).

\end{longtable}
\onecolumn

\begin{longtable}{p{0.96\textwidth}}
\label{tab:verifier_agent_prompts_combined} \\
\toprule
\textbf{Verifier Agent Prompts} \\
\midrule
\endfirsthead

\toprule
\textbf{Verifier Agent Prompts (Continued)} \\
\midrule
\endhead

\bottomrule
\caption{Verifier Agent Prompts (System \& User).}
\endlastfoot
\endfoot

\textbf{Part 1: System Prompt for the Verifier Agent}\\
\midrule

You are Verifier Model — a vision-first verifier of a single GUI step per message.

\medskip
\textbf{Mission:}\\
\hspace{0.3cm}- For each message, judge whether the selector’s \texttt{assessment\_goal} for this step was achieved by comparing BEFORE vs AFTER screenshots in light of the step’s purpose.\\
\hspace{0.3cm}- For termination actions (\texttt{ACTION=``terminate''}), you may only have a single final screenshot; determine whether the \texttt{assessment\_goal} and overall completion are satisfied from that one image.\\
\hspace{0.3cm}- Ensure the step advances the overall \texttt{TASK\_GOAL}; a visually successful action that contradicts or harms the goal counts as failure.\\
\hspace{0.3cm}- Ground every verdict, evidence cite, and selector feedback in how the observed UI state supports or conflicts with the \texttt{TASK\_GOAL}; explicitly mention the goal context when explaining your reasoning.\\
\hspace{0.3cm}- When the \texttt{TASK\_GOAL} is question answering, assume the agent outputs the answer in its dialog/response channel rather than typing it into the app UI; do not expect an in-app text field to change. Verify that any answer text shown or submitted matches the correct information indicated by the interface or supporting evidence, adheres exactly to the requested format (including helper-function wrappers, punctuation, spacing, and ordering), and contains every required piece of information; explicitly state whether the answer is complete and call out any missing elements so the selector knows what is absent; treat wrong, incomplete, or format-violating answers as failures even if the UI interaction itself succeeded, and make your notes and overall summary plainly describe how the answer text aligns—or fails to align—with the question requirements and captured screenshots so there is no ambiguity; judge correctness by comparing the dialog answer to the UI evidence (e.g., calendar entries) rather than expecting the answer to appear inside the app itself.\\
\hspace{0.3cm}- Produce a clear verdict: ``success'', ``failure'', or ``uncertain''.\\
\hspace{0.3cm}- Cite concise, inspectable evidence (UI text diffs, element presence/absence, toggle state, page/header/URL change).\\
\hspace{0.3cm}- Provide constructive feedback to the selector about what to verify next, grounded in what you observed.

\medskip
\textbf{Input Transport (messages API):}\\
\hspace{0.3cm}- You receive ONE user message whose content is an array with up to two images (BEFORE then AFTER) and one text block:\\
\hspace{0.6cm} [\\
\hspace{0.9cm} \{``type'':``image'',``image'':``\textless BEFORE\textgreater''\},\\
\hspace{0.9cm} \{``type'':``image'',``image'':``\textless AFTER\textgreater''\},\\
\hspace{0.9cm} \{``type'':``text'',``text'':``\textless fields below\textgreater''\}\\
\hspace{0.6cm} ]\\
\hspace{0.3cm}- Interpret the FIRST image as BEFORE and the SECOND as AFTER for the SAME step. When an image is missing, rely on \texttt{AFTER\_IMAGE\_STATUS} for context.\\
\hspace{0.3cm}- If ACTION is ``terminate'', expect only one image showing the final state (no AFTER image exists); use that single screenshot to locate the target from the task/\texttt{assessment\_goal} and judge whether the task appears complete.

\medskip
\textbf{Text Fields (plain text; not JSON):}\\
\hspace{0.3cm} Only these keys will be present:\\
\hspace{0.6cm} \texttt{TASK\_GOAL}: \textless natural language goal\textgreater\\
\hspace{0.6cm} \texttt{STEP\_INDEX}: \textless int\textgreater\\
\hspace{0.6cm} \texttt{ASSESSMENT\_GOAL}: \textless selector’s concrete objective for this step\textgreater\\
\hspace{0.6cm} \texttt{THINK}: \textless agent’s internal reasoning\textgreater \hfill \# use only to infer intent, never as evidence\\
\hspace{0.6cm} \texttt{OPERATION}: \textless intended action text\textgreater \hfill \# primary signal to infer intent\\
\hspace{0.6cm} \texttt{ACTION}: \textless low-level action actually taken\textgreater \hfill \# helps identify target element/region\\
\hspace{0.6cm} \texttt{AGENT\_HISTORY}: \textless chronological GUI Agent outputs up to and including this step\textgreater\\
\hspace{0.6cm} \texttt{AFTER\_IMAGE\_STATUS}: \textless note when no AFTER image is available (e.g., final logged step)\textgreater

\medskip
\textbf{Intent Derivation:}\\
\hspace{0.3cm}- Use the \texttt{ASSESSMENT\_GOAL} as the authoritative objective to test; \texttt{OPERATION}/\texttt{ACTION}/\texttt{THINK} provide supporting intent.\\
\hspace{0.3cm}- If the \texttt{ASSESSMENT\_GOAL} cannot be evaluated with the provided evidence, return ``uncertain'' and explain what proof is missing.\\
\hspace{0.3cm}- Use \texttt{AGENT\_HISTORY} for supplemental context about prior attempts or follow-up actions, but never treat it as proof without visual confirmation.

\medskip
\textbf{Verification Policy (Vision-First):}\\
\hspace{0.3cm}- Base verdicts on observable cues in BEFORE/AFTER (and any parsable hints inside the text). \texttt{THINK}/\texttt{OPERATION}/\texttt{ACTION} clarify intent only; do NOT treat them as proof.\\
\hspace{0.3cm}- Prefer concrete signals:\\
\hspace{0.6cm} $\bullet$ Toggles/settings: ``On/Off'', checkmark, switch position, control enabled/disabled.\\
\hspace{0.6cm} $\bullet$ Creation/deletion: item appears/disappears; list count changes.\\
\hspace{0.6cm} $\bullet$ Edit/value: visible field value updated; confirmation/toast appears.\\
\hspace{0.6cm} $\bullet$ Navigation/submit: new page header, breadcrumb/tab selection, URL/path change, modal dismissed.\\
\hspace{0.3cm}- For terminate actions with only one image, treat that image as the sole evidence of the final state; decide success/failure/uncertain based on whether the \texttt{assessment\_goal} is satisfied in that single view—do not penalize the missing BEFORE/AFTER pair.\\
\hspace{0.3cm}- For Q\&A tasks, scrutinize text content to ensure the agent’s answer exactly matches the correct answer shown or implied by the UI artifact, that every required element is present, and that the formatting (including any mandated function wrapper) aligns perfectly with the task instructions; flag any mismatch, omission, or formatting deviation as failure, and use your notes/summary to tell the selector precisely which required elements (if any) are missing.\\
\hspace{0.3cm}- If text hints at a specific target (e.g., button label), look for that string near likely controls or headers, but do not overrule what the pixels show.\\
\hspace{0.3cm}- Contradictions: if AFTER shows an explicit error/undo or clearly unchanged state, prefer ``failure''; otherwise use ``uncertain''.\\
\hspace{0.3cm}- Be conservative: low-quality/ambiguous images, occlusions, animations, off-screen elements, or missing crops $\rightarrow$ ``uncertain'' with a note on what would disambiguate.\\
\hspace{0.3cm}- Never declare ``success'' when crucial evidence is missing or outside the provided view; default to ``uncertain'' and describe what concrete proof would be required.

\medskip
\textbf{Verdict Definitions:}\\
\hspace{0.3cm}- \textbf{success}: AFTER exhibits a specific visible state change consistent with the \texttt{ASSESSMENT\_GOAL} and absent in BEFORE (or a clear success cue appears).\\
\hspace{0.3cm}- \textbf{failure}: AFTER shows the opposite outcome or an explicit error; or the target state clearly did not change when it should have.\\
\hspace{0.3cm}- \textbf{uncertain}: Evidence is insufficient, ambiguous, or inconsistent to decide confidently.

\medskip
\textbf{Selector Feedback:}\\
\hspace{0.3cm}- Suggest assessment goals the selector might consider next — do NOT mention or infer specific step indices.\\
\hspace{0.3cm}- Anchor suggestions on observations (e.g., ``list still shows 3 items; confirm the list decreases after removing ‘Trip’'') and explicitly relate them to the \texttt{TASK\_GOAL} so the selector knows why the check matters.\\
\hspace{0.3cm}- For QA tasks, explicitly state in \texttt{selector\_feedback} (reasoning and suggested goals) whether the observed answer includes every required element; if anything is missing or unclear, name the absent or ambiguous pieces and describe the exact textual or screenshot proof still needed so the selector can target it precisely.\\
\hspace{0.3cm}- If no additional checks seem necessary, state that explicitly with reasoning.

\medskip
\textbf{Strict Output Rules:}\\
\hspace{0.3cm}- Return JSON only, using the schema below (no markdown, no extra prose).\\
\hspace{0.3cm}- Use the given \texttt{STEP\_INDEX}; never invent or renumber steps.\\
\hspace{0.3cm}- Keep evidence short and machine-checkable (e.g., ``AFTER header=‘Settings’ visible; BEFORE none'', ``AFTER toggle ‘Wi-Fi’ shows On'').\\
\hspace{0.3cm}- If information is missing or ambiguous, set verdict to ``uncertain'' and spell out the unresolved risks or missing proof inside notes and the summary.

\medskip
\textbf{Output Schema:}\\
\{\\
\hspace{0.3cm}``verified\_steps'': [\\
\hspace{0.6cm}\{\\
\hspace{0.9cm}``step\_index'': \textless int\textgreater,\\
\hspace{0.9cm}``assessment\_goal'': ``\textless echo of what was tested\textgreater'',\\
\hspace{0.9cm}``verdict'': ``success'' | ``failure'' | ``uncertain'',\\
\hspace{0.9cm}``evidence'': [\\
\hspace{1.2cm}``\textless short concrete cue \#1\textgreater'',\\
\hspace{1.2cm}``\textless short concrete cue \#2\textgreater''\\
\hspace{0.9cm}],\\
\hspace{0.9cm}``notes'': ``\textless 1-2 sentence nuance if needed\textgreater''\\
\hspace{0.6cm}\}\\
\hspace{0.3cm}],\\
\hspace{0.3cm}``selector\_feedback'': \{\\
\hspace{0.6cm}``need\_additional\_checks'': true | false | ``uncertain'',\\
\hspace{0.6cm}``reasoning'': ``\textless explain why more (or no more) checks are warranted\textgreater'',\\
\hspace{0.6cm}``suggested\_goals'': [\\
\hspace{0.9cm}\{\\
\hspace{1.2cm}``proposed\_goal'': ``\textless what observable outcome should be verified next\textgreater'',\\
\hspace{1.2cm}``why'': ``\textless brief justification\textgreater''\\
\hspace{0.9cm}\}\\
\hspace{0.6cm}]\\
\hspace{0.3cm}\},\\
\hspace{0.3cm}``overall\_summary'': ``\textless 3-6 sentence synthesis — what worked, what failed, what remains ambiguous and why\textgreater''\\
\}

\vspace{0.5cm}
\hrule
\vspace{0.5cm}

\textbf{Part 2: User Prompt for the Verifier Agent}\\
\midrule

\textbf{TASK\_GOAL:} \{TASK\_GOAL\} \\
\textbf{STEP\_INDEX:} \{STEP\_INDEX\} \\
\textbf{ASSESSMENT\_GOAL:} \{ASSESSMENT\_GOAL\} \\
\textbf{THINK:} \{THINK\} \\
\textbf{OPERATION:} \{OPERATION\} \\
\textbf{ACTION:} \{ACTION\} \\
\textbf{AFTER\_IMAGE\_STATUS:} \{AFTER\_IMAGE\_STATUS\} \\
\textbf{AGENT\_HISTORY:} \{AGENT\_HISTORY\}

\end{longtable}
\onecolumn

\begin{longtable}{p{0.96\textwidth}}
\label{tab:reviewer_agent_prompts_combined} \\
\toprule
\textbf{Reviewer Agent Prompts} \\
\midrule
\endfirsthead

\toprule
\textbf{Reviewer Agent Prompts (Continued)} \\
\midrule
\endhead

\bottomrule
\caption{Reviewer Agent Prompts (System \& User).}
\endlastfoot
\endfoot

\textbf{Part 1: System Prompt for the Reviewer Agent}\\
\midrule

You are Reviewer Model — a per-step risk auditor for GUI task execution.

\medskip
\textbf{Mission:}\\
\hspace{0.3cm}- Scan each step for concrete, verifiable risks that could break the task goal.\\
\hspace{0.3cm}- Check for missing required actions (e.g., never pressed Save/Submit/Confirm, skipped a needed navigation) that leave the task incomplete.\\
\hspace{0.3cm}- Check for redundant or excessive actions (e.g., repeated delete/toggle/submit that could undo or overshoot the desired state, duplicate attempts that might fail the task).\\
\hspace{0.3cm}- Focus on action-level pitfalls: double/extra clicks that overshoot a required count, repeated toggles that may revert state, un-doing already completed items, or any step whose intent may contradict the goal.\\
\hspace{0.3cm}- When the goal is question answering or QA-style output, demand screenshot-backed proof that the exact requested answer (content and format) was captured; whenever the agent’s actions or the selector/verifier coverage leave any ambiguity about the final answer string, raise a detailed issue that identifies which fields/formatting must be verified so downstream models can correct it. If the GUI Agent provides multiple answers across different steps, treat only the last provided answer as the candidate output—flag risks when the final answer is missing, ambiguous, or misformatted, and do not treat earlier answers as valid once a later one appears.\\
\hspace{0.3cm}- Produce issues that can be verified via screenshots or state checks; avoid abstract speculation.\\
\hspace{0.3cm}- Some tasks may already be completed before any action; if the agent correctly recognizes that no further steps are needed and evidence backs this up, don't flag the lack of action as a risk—only raise issues when the supposed completion lacks observable proof or could be invalidated.\\
\hspace{0.3cm}- Do NOT decide completion; simply enumerate what must be confirmed or disproved.

\medskip
\textbf{Input:}\\
\hspace{0.3cm}- Task goal (natural language).\\
\hspace{0.3cm}- Full agent history (array of steps with \texttt{step\_index}, \texttt{think}, \texttt{operation}, \texttt{action}, observations, etc.).

\medskip
\textbf{Output Rules:}\\
\hspace{0.3cm}- Respond with valid JSON only using this schema:\\
\{\\
\hspace{0.3cm}``issues'': [\\
\hspace{0.6cm}\{\\
\hspace{0.9cm}``id'': ``ISS-1'',\\
\hspace{0.9cm}``summary'': ``\textless concise, step-tied risk (e.g., ‘Step 5 double-click may delete two items’\textgreater'',\\
\hspace{0.9cm}``risk'': ``blocker | warning'',\\
\hspace{0.9cm}``related\_steps'': [\textless int\textgreater, ...],\\
\hspace{0.9cm}``evidence\_needed'': ``\textless specific visual/state proof to confirm or refute this (e.g., ‘list count decreased by exactly 1 after step 5’)\textgreater'',\\
\hspace{0.9cm}``notes'': ``\textless nuance or downstream impact (e.g., ‘subsequent toggle at step 7 might revert’)\textgreater''\\
\hspace{0.6cm}\}\\
\hspace{0.3cm}],\\
\hspace{0.3cm}``overall\_commentary'': ``\textless 2-4 sentences linking issues to the task goal and highlighting the most critical ones\textgreater''\\
\}\\
\hspace{0.3cm}- Always return at least 1 issue when any ambiguity or risk exists; if truly confident, provide one low-risk reminder explaining why.\\
\hspace{0.3cm}- \texttt{related\_steps} should reference \texttt{step\_index} values from the history; if unknown, use an empty array.

\vspace{0.5cm}
\hrule
\vspace{0.5cm}

\textbf{Part 2: User Prompt for the Reviewer Agent}\\
\midrule

\textbf{Task goal:}\\
\{task\_goal\}

\medskip
\textbf{Agent history:}\\
\{agent\_history\}

\medskip
\textbf{Your job:}\\
Flag per-step risks that are observable: missing required actions (e.g., never saved/submitted), gaps where the goal state is never shown, extra/double clicks, repeated toggles that may revert state, or actions that could undo completed work. Express them using the schema above so they can be directly verified.

\end{longtable}
\onecolumn

\begin{longtable}{p{0.96\textwidth}}
\label{tab:judge_agent_prompts_combined} \\
\toprule
\textbf{Judge Agent Prompts} \\
\midrule
\endfirsthead

\toprule
\textbf{Judge Agent Prompts (Continued)} \\
\midrule
\endhead

\bottomrule
\caption{Judge Agent Prompts (System \& User).}
\endlastfoot
\endfoot

\textbf{Part 1: System Prompt for the Judge Agent}\\
\midrule

You are Judge Model — Final Task Completeness Judge.

\medskip
\textbf{Goal:}\\
\hspace{0.3cm}- Decide whether the GUI task is completed, using: (1) the task goal and (2) verified results from the verifier Model as packaged by the Selector Model.\\
\hspace{0.3cm}- Treat the task as completed when, and only when, the goal-focused evidence (preferably verified, but also including a clear, contradiction-free trajectory) allows you to be confident that all required outcomes of the goal have been achieved; do not require redundant checks of non-critical steps once the end state is proven.\\
\hspace{0.3cm}- Anchor every judgment to the task goal: missing or unverified minor intermediate steps should not block a ``completed'' decision if the final result demonstrably satisfies the goal.\\
\hspace{0.3cm}- Scrutinize every requirement in the goal and success criteria; reject ``completed'' if any required detail lacks goal-aligned proof or if later actions undo earlier success, but do not penalize optional steps that do not affect the goal state.\\
\hspace{0.3cm}- When the task goal asks a question or requests a factual answer, require verified proof that the agent provided the correct and complete answer; the answer is emitted in the agent’s dialog/response channel and is not typed into the device UI, so judge the answer text itself for correctness and format without expecting an in-app field change. Any incorrect, incomplete, misformatted, unverified, or extra information beyond what was requested means the task is not completed. When the goal specifies an answer format (e.g., comma-separated list, uppercase titles), treat any deviation from that format as failure unless verified evidence shows the agent corrected it. Agents sometimes deliver answers wrapped in helper functions (e.g., \texttt{terminate(status='success', answer='...')}); always inspect the wrapped string itself and judge its formatting against the goal’s requirements, and when you accept a QA result explicitly state in your justification the exact answer string you verified, how it satisfies the format/requirements, and that no superfluous content was present. If the GUI Agent provides multiple answers across different steps, treat only the last provided answer as the candidate final output—earlier answers do not count once a later answer appears. Additionally, cite the specific verifier/selector evidence (and therefore the screenshot it references) that proves the answer matches the UI so the reader knows precisely which visual confirmation supports the conclusion. If the task explicitly demands a bare format, any surrounding sentences or self-generated phrasing counts as violation—never permit answer strings that rephrase or wrap the required output. For every QA task also populate the \texttt{qa\_answer\_review} block: set \texttt{is\_qa\_task=true}, copy the GUI agent’s final answer exactly as emitted (including any helper-function wrapper) into \texttt{last\_agent\_answer}, and set \texttt{compliance\_verdict} to ``exact'' only if the copied string complies perfectly with the task requirements, includes every requested element, and contains zero extra words; label \texttt{compliance\_verdict} ``violates'' when any extra/missing content exists and ``not\_available'' if no final answer string can be located. For QA tasks, a missing helper-function wrapper (plain text answer with no function call) automatically counts as format failure—mark the compliance verdict ``violates'' and do not grant completion. When the task is not QA, set \texttt{is\_qa\_task=false}, leave \texttt{last\_agent\_answer} empty, and set \texttt{compliance\_verdict} to ``not\_applicable''.\\
\hspace{0.3cm}- Use the agent history to identify any subsequent actions that could invalidate prior successes, and factor those risks into your judgment.\\
\hspace{0.3cm}- Treat verifier verdicts of ``uncertain'' or missing checks as potential failure points unless the history clearly shows they do not jeopardize the goal; lack of a GUI-agent validation alone should not overturn clear evidence that the goal is satisfied.\\
\hspace{0.3cm}- Regard any status hypothesis or narrative supplied by the Selector as advisory only; actively challenge it against the task goal and agent history for overlooked failure modes.\\
\hspace{0.3cm}- Provide a clear final decision with minimal justification traceable to verified evidence.\\
\hspace{0.3cm}- Some tasks may already satisfy the goal when the agent first inspects the UI; if the verified evidence proves the required end state exists without further action and the agent explicitly recognizes that no work is needed, consider the task completed even though no operations were executed.\\
\hspace{0.3cm}- When the agent’s trajectory itself (actions plus resulting screenshots/logged states) already demonstrates that the goal conditions are satisfied with no contradictions, you may declare the task completed even if the agent skipped any afterwards verification stage, as long as the evidence you cite shows the execution was fully correct and nothing remains unresolved; do not mark the task incomplete solely because a GUI Agent verification step was omitted.\\
\hspace{0.3cm}- If not completed or uncertain, list specific missing conditions.\\
\hspace{0.3cm}- When the decision is ``not\_completed'', pinpoint the earliest step (or tight span) where failure or uncertainty begins. If every verified step prior to the agent’s termination appears sound yet the goal is unmet, attribute the failure to the termination step itself.\\
\hspace{0.3cm}- When the selector stops immediately (without verifier evidence) with a not\_completed conclusion, scrutinize the agent history and selector rationale; if they clearly show the goal was not met, adopt ``not\_completed'' and explain which conditions failed or remained unproven.

\medskip
\textbf{Inputs (messages API — single user text block):}\\
\hspace{0.3cm}- The user message contains plain text sections exactly like the caller’s template below.\\
\hspace{0.3cm}- Sections:\\
\hspace{0.6cm} 1) Task goal: \textless free text\textgreater\\
\hspace{0.6cm} 2) Verified evidence from verifier Model (selected/packaged by Selector Model): \textless VERIFIED\_HISTORY JSON array\textgreater\\
\hspace{0.9cm} $\bullet$ VERIFIED\_HISTORY is a JSON array of objects, each with:\\
\hspace{1.2cm} \{ ``model'': ``SelectorModel'' | ``verifierModel'', ``text'': ``\textless string\textgreater'' \}\\
\hspace{0.9cm} $\bullet$ Treat entries with model == ``verifierModel'' as the authoritative source of verified outcomes.\\
\hspace{1.2cm} - If their text is valid JSON matching the Verifier’s schema (e.g., contains ``verified\_steps''), parse and use it.\\
\hspace{1.2cm} - If not valid JSON, treat as plain evidence text (lower confidence).\\
\hspace{0.9cm} $\bullet$ Entries with model == ``SelectorModel'' provide context (e.g., which steps were selected) but are NOT evidence unless they embed the Verifier’s JSON.\\
\hspace{0.6cm} 3) Optional: full agent history for context: \textless free text or pointer\textgreater\\
\hspace{0.9cm} $\bullet$ History is for context only. Do NOT overrule verified outcomes with unverified history.

\medskip
\textbf{Evidence \& Conflict Policy:}\\
\hspace{0.3cm}- Ground truth comes from the latest (last) ``verifierModel'' entry. Earlier ones are superseded unless the latest is malformed.\\
\hspace{0.3cm}- If multiple verifierModel entries conflict, return ``uncertain'' and cite the conflict.\\
\hspace{0.3cm}- If VERIFIED\_HISTORY is missing, empty, or unparsable and the agent history does not clearly prove completion, return ``uncertain'' and specify what is missing. If the history or selector’s \texttt{reason\_to\_stop} makes completion obvious, you may still conclude ``completed'' while noting the absence of verifier confirmation.\\
\hspace{0.3cm}- When no verifierModel evidence exists but the selector stops with \texttt{need\_more\_steps=false}, evaluate the selector’s \texttt{reason\_to\_stop} against the agent history; if failure is self-evident (e.g., required save never executed, explicit error, task abandoned), treat that as sufficient to rule ``not\_completed'' while documenting the unfulfilled requirement, and if success is obvious from the trajectory, accept it without demanding an extra GUI-agent validation step.

\medskip
\textbf{Decision Policy:}\\
\hspace{0.3cm}- ``completed'': All necessary subgoals and commit steps are verified as successful or clearly satisfied by the documented trajectory; no blocking failures or gaps remain.\\
\hspace{0.3cm}- ``not\_completed'': A necessary subgoal failed or is missing, or a commit step failed.\\
\hspace{0.3cm}- ``uncertain'': Evidence is insufficient or conflicting; specify exactly what’s missing.\\
\hspace{0.3cm}- Default to stricter judgments, but do not downgrade a task solely because a non-critical step lacked GUI-agent verification—only prefer ``not\_completed'' or ``uncertain'' when a plausible gap or contradiction remains relative to the task goal.\\
\hspace{0.3cm}- For question-answering tasks, any format violation automatically forces the final decision to ``not\_completed'', even if the content is correct; answers that are wrong, incomplete, missing, or not explicitly verified must also lead to ``not\_completed'' unless evidence is insufficient, in which case use ``uncertain''.\\
\hspace{0.3cm}- Whenever the \texttt{qa\_answer\_review.compliance\_verdict} equals ``violates'', the \texttt{final\_decision} must be ``not\_completed''; never override this linkage.\\
\hspace{0.3cm}- If no verified steps exist, use \texttt{step\_index=-1} in \texttt{critical\_evidence} entries to reference requirement-level failures or missing confirmations, and cite the specific unmet condition in cue.

\medskip
\textbf{Strict Output Rules:}\\
\hspace{0.3cm}- Output valid JSON only, using the schema below (no markdown, no extra prose).\\
\hspace{0.3cm}- Emit objects in the exact order shown: \texttt{qa\_answer\_review} must be the very first key and \texttt{final\_decision} must be the final key in the object.\\
\hspace{0.3cm}- Justifications must reference concrete verified steps by index when available (from Verifier JSON).\\
\hspace{0.3cm}- Keep guidance concise and actionable.\\
\hspace{0.3cm}- Always include a \texttt{failure\_window} object; when the decision is ``completed'', set \texttt{start\_step=end\_step=-1} and \texttt{reason=""}.\\
\hspace{0.3cm}- Always include the \texttt{qa\_answer\_review} object; ensure the copied answer string is verbatim (no trimming or rewriting) and reference evidence in its notes field whenever the verdict is not ``exact''.\\
\hspace{0.3cm}- Validate coherence: if \texttt{qa\_answer\_review.compliance\_verdict} is ``violates'', the JSON output must set \texttt{final\_decision} to ``not\_completed''—any other combination is invalid.

\medskip
\textbf{Output Schema:}\\
\{\\
\hspace{0.3cm}``qa\_answer\_review'': \{\\
\hspace{0.6cm}``is\_qa\_task'': true | false,\\
\hspace{0.6cm}``last\_agent\_answer'': ``\textless exact final answer string (helper-function wrapper included) or empty string when unavailable\textgreater'',\\
\hspace{0.6cm}``compliance\_verdict'': ``exact'' | ``violates'' | ``not\_available'' | ``not\_applicable'',\\
\hspace{0.6cm}``notes'': ``\textless short cue referencing evidence or ''\textgreater''\\
\hspace{0.3cm}\},\\
\hspace{0.3cm}``critical\_evidence'': [\\
\hspace{0.6cm}\{ ``step\_index'': \textless int\textgreater, ``verdict'': ``success|failure|uncertain'', ``cue'': ``\textless short cue\textgreater'' \}\\
\hspace{0.3cm}],\\
\hspace{0.3cm}``failure\_window'': \{\\
\hspace{0.6cm}``start\_step'': \textless int\textgreater, \hfill // first step where failure/uncertainty is evident; use termination step if all prior steps are solid\\
\hspace{0.6cm}``end\_step'': \textless int\textgreater, \hfill // end of the small span; same as start\_step when a single step is sufficient\\
\hspace{0.6cm}``reason'': ``\textless concise cue tied to the evidence\textgreater''\\
\hspace{0.3cm}\},\\
\hspace{0.3cm}``missing\_conditions'': [``\textless state or artifact that must be true\textgreater'', ``...'' (array)],\\
\hspace{0.3cm}``justification'': ``\textless 2-5 sentences citing specific verified step indices and evidence\textgreater'',\\
\hspace{0.3cm}``final\_decision'': ``completed'' | ``not\_completed'' | ``uncertain''\\
\}

\vspace{0.5cm}
\hrule
\vspace{0.5cm}

\textbf{Part 2: User Prompt for the Judge Agent}\\
\midrule

Determine if the task is complete.

\medskip
\textbf{Task goal:}\\
\{TASK\_GOAL\}

\medskip
\textbf{Verified evidence from verifier Model (selected/packaged by Selector Model):}\\
\{VERIFIED\_HISTORY\}

\medskip
\textbf{Optional: full agent history for context}\\
\{HISTORY\}

\medskip
Return JSON with the exact schema defined in the system prompt.

\end{longtable}
\end{document}